\documentclass[lettersize,journal]{IEEEtran}
\usepackage{algorithm}
\usepackage{algpseudocode}  

\usepackage{array}
\usepackage[caption=false,font=normalsize,labelfont=sf,textfont=sf]{subfig}
\usepackage{textcomp}
\usepackage{stfloats}
\usepackage{url}
\usepackage{verbatim}
\usepackage{graphicx}
\usepackage{cite}
\usepackage{caption}
\usepackage{graphics} 
\usepackage{epsfig} 

\usepackage{cuted}
\usepackage{newtxtext}
\usepackage{newtxmath}
\usepackage{amsmath} 
\usepackage{makecell}
\usepackage{booktabs}
\usepackage{multirow}
\usepackage{acronym}
\usepackage{hyperref}
\usepackage{xurl}
\usepackage{xcolor}
\usepackage{soul}
\usepackage[normalem]{ulem}
\usepackage{tikz}
\usepackage[labelfont=bf]{caption}

\acrodef{DVS}[DVS]{Dynamic Vision Sensor}
\acrodef{RL}[RL]{reinforcement learning}
\acrodef{ROI}[ROI]{Region of Interest}
\acrodef{DDPG}[DDPG]{Deep Deterministic Policy Gradient}
\acrodef{MLP}[MLP]{Multilayer Perceptron}
\acrodef{DBSCAN}[DBSCAN]{Density-Based Spatial Clustering}
\acrodef{CDTA}[CDTA]{case-dependent temporally-adaptive}
\acrodef{SOTA}[SOTA]{state-of-the-art}

\definecolor{RGBColor}{HTML}{AABCDB}  
\definecolor{DVSColor}{HTML}{487db2}  

\definecolor{fusionColor}{HTML}{F7D4DB}  
\definecolor{EBPPColor}{HTML}{E595A4}  
\definecolor{oursColor}{HTML}{CE4459}  

\usepackage{subfig}
\begin{document}

\title{Biologically Inspired Event-Based Perception and Sample-Efficient Learning for High-Speed Table Tennis Robots}

\author{Ziqi Wang$^{1}$, Jingyue Zhao$^{2}$, Xun Xiao$^{1}$, Jichao Yang$^{3}$, Yaohua Wang$^{1}$\\ Shi Xu$^{2}$, Lei Wang$^{2,3}$ and Huadong Dai$^{2}$%
\thanks{This work was supported in part by the National Natural Science Foundation of China under Grants 62372461,(Corresponding author: Lei Wang)}
\thanks{$^{1}$Ziqi Wang, Xun Xiao and Yaohua Wang are with National University of Defence Technology, Changsha 410073, Hunan, P.R.China. E-mail:
        {\tt\small wangziqi@nudt.edu.cn}}%
\thanks{$^{2}$Jingyue Zhao, Shi Xu, Lei Wang and Huadong Dai are with Defense Innovation Institute, AMS. E-mail:
        {\tt\small leiwang@nudt.edu.cn}}%
\thanks{$^{3}$Jichao Yang and Lei Wang is with Qiyuan Lab}
}

\maketitle

\begin{abstract}
Perception and decision-making in high-speed dynamic scenarios remain challenging for current robots. 
In contrast, humans and animals can rapidly perceive and make decisions in such environments.
Taking table tennis as a typical example, conventional frame-based vision sensors suffer from motion blur, high latency and data redundancy, which can hardly meet real-time, accurate perception requirements.
Inspired by the human visual system, event-based perception methods address these limitations through asynchronous sensing, high temporal resolution, and inherently sparse data representations. However, current event-based methods are still restricted to simplified,  unrealistic ball-only scenarios.
Meanwhile, existing decision-making approaches typically require thousands of interactions with the environment to converge, resulting in significant computational costs.
In this work, we present a biologically inspired approach for high-speed table tennis robots, combining event-based perception with sample-efficient learning.
On the perception side, we propose an event-based ball detection method that leverages motion cues and geometric consistency, operating directly on asynchronous event streams without frame reconstruction, to achieve robust and efficient detection in real-world rallies. This approach achieves up to 99.8\% reduction in processed data, 96.4\% reduction in latency, and 27.5\% improvement in recall over frame-based methods, while achieving up to a 89.7\% accuracy gain over the existing event-based approaches. 
On the decision-making side, we introduce a human-inspired, sample-efficient training strategy that first trains policies in low-speed scenarios, progressively acquiring skills from basic to advanced, and then adapts them to high-speed scenarios, guided by a case-dependent temporally-adaptive reward and a reward-threshold mechanism. With the same training episodes, our method improves return-to-target accuracy by 35.8\%.
These results demonstrate the effectiveness of biologically inspired perception and decision-making for high-speed robotic systems.

\end{abstract}


\section{Introduction}

Fast and reliable perception and decision-making in high-speed dynamic environments remains a significant challenge for modern robots, because of rapidly-moving targets, limited reaction time, and stringent tolerance to errors~\cite{falanga2020dynamic,he2021fast}.
Table tennis exemplifies such environments, as the ball exhibits a set of complex characteristics, including small size, high velocity, and frequent bouncing.
Moreover, the extremely limited temporal window for striking the ball makes the robot highly sensitive to even minor decision errors, which can be rapidly amplified and severely degrade shot performance.
These unique motion and interaction properties make table tennis an ideal testbed for studying perception and decision-making in high-speed dynamic environments~\cite{buchler2022learning,dambrosio2025achieving}. 

In recent years, many studies have made notable progress in improving the perception and decision-making performance of table tennis robots by employing conventional frame-based cameras for perception and \ac{RL} for hitting strategy learning~\cite{d2023robotic,cursi2024safe}. However, two crucial challenges still retard the development of high-speed table tennis robots.
\textbf{First, how to reliably acquire the real-time positions of a fast-moving table tennis ball.}
Frame-based sensing and processing pipelines introduce substantial latency, which directly limits the time available for decision-making~\cite{chen2025low}. Moreover, fast-moving targets are prone to severe motion blur, which degrades perception accuracy~\cite{ziegler2025detection}. In addition, frame-based imaging generates highly redundant data, imposing considerable storage and computational overhead and ultimately constraining the overall system efficiency~\cite{lichtsteiner2008128}.
\textbf{Second, how to achieve sample-efficient learning of hitting strategies.} Existing \ac{RL} methods typically rely on extensive environment interactions, which incur prohibitive costs in both real-world settings and simulation~\cite{huang2016jointly,henderson2018deep}. Moreover, prolonged training in simulation may introduce non-negligible simulation-to-reality discrepancies, particularly in highly dynamic scenarios, thereby making sample efficiency a critical consideration~\cite{peng2018sim}. In contrast, humans exhibit remarkable table tennis playing abilities, enabling them to rapidly and accurately perceive the ball position in highly dynamic environments and to efficiently acquire effective hitting strategies.

\begin{figure*}[t]
  \centering
  \includegraphics[width=1\textwidth]{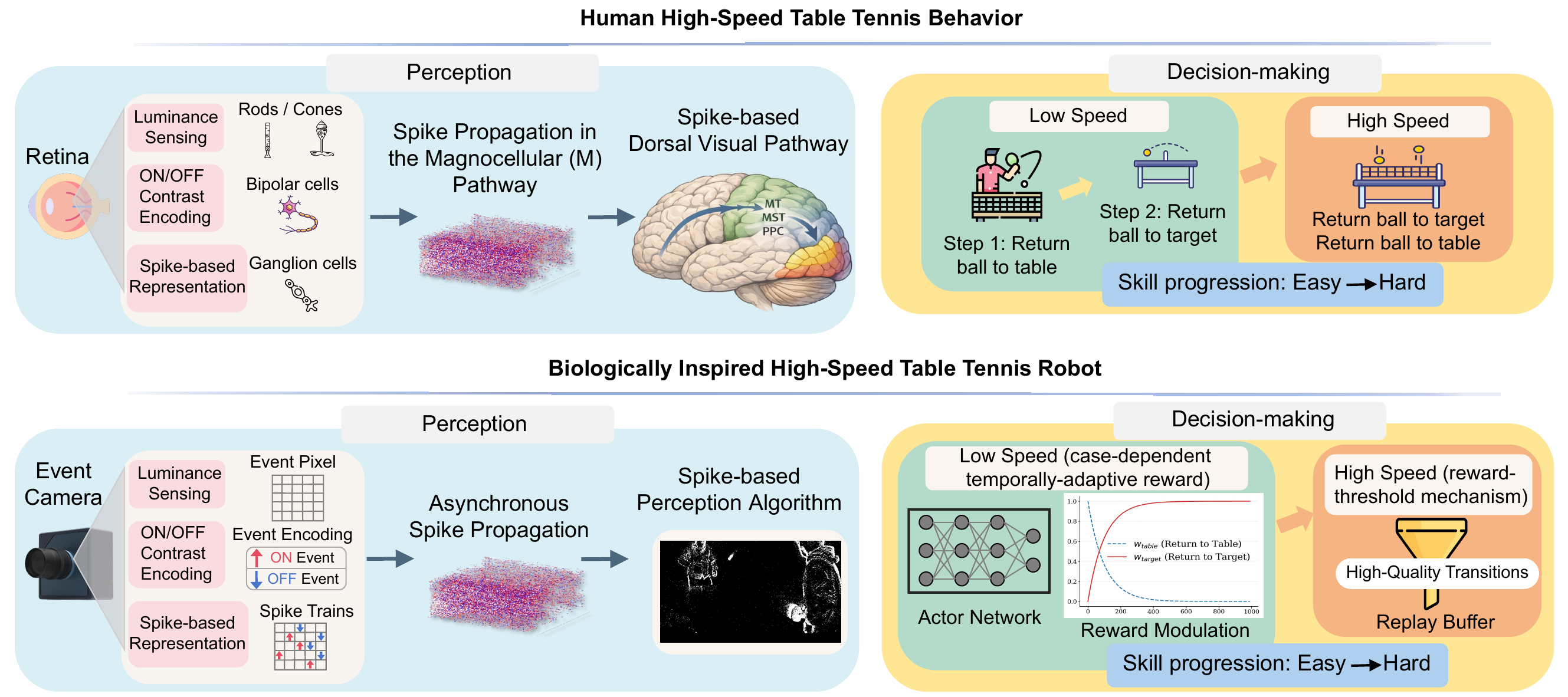}
    \captionof{figure}{\textbf{The mechanism of table tennis in humans and robots.} Humans perceive the position of the ball through the retina, transmitting the information as spikes via the M pathway to the dorsal visual stream to support motion perception. During the process of learning table tennis skills, humans typically learn table tennis skills in a progression from easy to hard. Inspired by this, robots can train in a brain-inspired manner. They use event cameras to achieve perception by processing asynchronous event streams. In terms of decision learning, robots first train in low-speed scenarios, gradually acquiring basic skills before progressing to more advanced capabilities to achieve policy convergence, and then continue training in high-speed scenarios.}
    \label{pic_brain}
\end{figure*}

In terms of perception (see the perception module in Fig.~\ref{pic_brain}), visual information in humans is initially encoded at the retina through rods and cones~\cite{baylor1979responses}, bipolar cells~\cite{schiller1992and}, and ganglion cells~\cite{kuffler1953discharge}, which collaboratively transform luminance variations into spike-based neural signals. These signals are predominantly conveyed by the magnocellular (M) pathway, which is highly sensitive to luminance contrast and high temporal-frequency changes~\cite{merigan1993parallel}. The M pathway further projects to the dorsal visual stream, where rapid, transient neural responses support motion perception and enable efficient processing of fast-moving objects~\cite{milner2008two}. Inspired by human visual processing, neuromorphic perception using \ac{DVS} has emerged as a promising solution~\cite{xiao2022dynamic,ziegler2025event}. \ac{DVS} employs an array of event pixels that asynchronously detect local brightness changes and convert them through event encoding into ON/OFF events. These events form sparse, high-temporal-resolution spike trains, which are then organized into an asynchronous event stream~\cite{gallego2020event}.
Consequently, motion blur is effectively alleviated, and interference from the static background is largely eliminated. Simultaneously, the event-driven mechanism enables rapid perception with minimal data throughput. 

Despite the great potential of \ac{DVS} for table-tennis perception, related studies are still in their early stages. In particular, they mainly assume idealized scenarios where only the ball is present within the visual scene, neglecting interference from player motions and other dynamic objects~\cite{ziegler2025event}. In actual gameplay, multiple moving objects and player actions significantly complicate target identification. To address this, we propose a robust event-based perception framework capable of operating reliably in real-world table tennis rallies, which emulates the dorsal visual pathway of the brain and processes asynchronous event streams for motion perception. Specifically, we leverage the spatiotemporal structure and polarity characteristics inherent in event streams, along with geometric constraints, to enhance the accuracy of ball localization for subsequent hitting decision-making.

In terms of decision-making (see the decision-making module in Fig.~\ref{pic_brain}), humans gradually acquire stable hitting strategies through practice. In this process, humans initially learn in low-speed scenarios, first mastering fundamental capabilities (e.g., returning the ball onto the table) and then progressing to more advanced skills (e.g., accurately returning the ball to a target location). Once proficiency is achieved at this speed, the ball velocity is gradually increased, and the acquired skills are further practiced under higher-speed conditions.~\cite{schmidt1975schema,beilock2001fragility}
Inspired by human skill acquisition, the policy is initially trained under low ball velocities to avoid convergence issues caused by excessively high speeds, and the ball speed is then progressively increased as the policy converges.
Despite its effectiveness, this learning process introduces two key challenges. One challenge is how to achieve sample-efficient learning across different velocity stages. 
Accordingly, we employ a \ac{CDTA} reward. Unlike prior approaches that use a fixed reward function throughout the entire training process, we dynamically adjust reward signals to guide the agent in smoothly progressing from simpler tasks to more complex ones. This approach significantly enhances the sample efficiency of policy learning, facilitating faster acquisition of effective hitting strategies.

Another challenge is how to effectively leverage the converged low-speed policy in higher-speed scenarios, particularly in terms of managing the replay buffer carried over from prior training.
Buffer discarding and retention are common replay strategies, which are limited by the need to re-collect transitions and interference from low-value transitions~\cite{fedus2020revisiting,isele2018selective}. 
To address this issue, we propose a reward-threshold mechanism that selectively retains only high-quality transitions for updates. Experimental results demonstrate that this mechanism effectively leverages prior experience to accelerate the learning of hitting strategies under more challenging high-speed conditions.

In this work, we report a biologically inspired high-speed table tennis robots. The robot emulates the information encoding and perception mechanisms of biological vision and draws inspiration from the human learning process from simple to complex tasks. As a result, it enables robust and efficient perception and decision-making in highly dynamic environments. We addressed several key challenges in the design.

\begin{itemize}
\item We present a high-accuracy, low-latency event-based perception method for table tennis in complex dynamic environments. Our approach achieves up to 99.8\% reduction in processed data, 96.4\% reduction in latency, and 27.5\% improvement in recall over frame-based methods, while achieving up to a 89.7\% accuracy gain over existing event-based approaches.

\item We propose a human-inspired, sample-efficient training strategy based on \ac{DDPG}, which combines a \ac{CDTA} reward with a reward-threshold mechanism, achieving at least 35.8\% improvement in return accuracy over the \ac{SOTA} method under the same number of training episodes.

\item To support future research, we release for the first time a high-quality real-world DVS table tennis dataset that closely reflects practical application scenarios.

\end{itemize}

\section{RELATED WORK}

\subsection{Perception}
\subsubsection{Frame-Based Methods}
Existing frame-based perception methods can be broadly categorized into model-based and learning-based approaches~\cite{zhang2024detecting,gomez2019reliable}. Early studies predominantly adopted model-based methods, which detect the ball by integrating motion cues, color features, and geometric constraints. For instance, \cite{li2012ping} proposed a multi-threshold ball detection method that combines frame differencing, HSV color space filtering, and geometric features such as circularity and area to improve detection accuracy. \cite{tebbe2018table} further enhanced robustness under high-frame-rate conditions by introducing non-adjacent frame differencing, while jointly leveraging color and geometric constraints to suppress background interference. Subsequent work \cite{tebbe2022adaptive} reduced computational complexity and processing latency by employing single-channel image differencing. In recent years, learning-based approaches have gained increasing attention. These methods typically rely on deep neural networks trained end-to-end on annotated datasets for table tennis ball detection. For example, \cite{wang2025spikepingpong} trained a YOLOv4-tiny model on two public datasets (TT2 and Ping Pong Detection). However, such approaches remain prone to false detections of static circular objects in complex scenes and exhibit a strong dependence on large-scale labeled data.

Overall, frame-based detection methods suffer from inherent limitations in high-speed scenarios, including motion blur, high data redundancy, and relatively large perception latency, which significantly constrain system responsiveness and stability. In contrast, event-based cameras offer high temporal resolution and sparse data representations, effectively alleviating these issues and providing more timely and reliable perceptual inputs for subsequent decision-making modules.

\begin{figure}[t]
  \centering
  \includegraphics[width=0.48\textwidth]{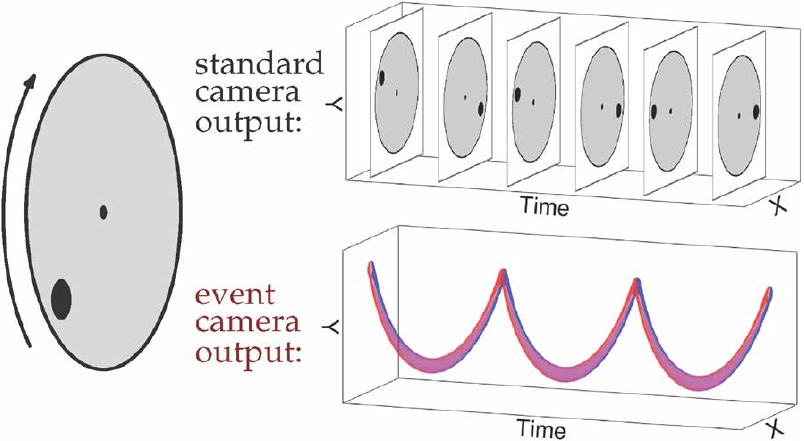}
  \caption{\textbf{Comparison between conventional cameras and event cameras.} 
When imaging a rotating disk with black dots, conventional cameras capture image frames at a fixed frame rate.
In contrast, event cameras record brightness changes in the form of an asynchronous event stream and generate events only when the intensity change exceeds a predefined threshold \cite{falanga2020dynamic}.
}
  \label{pic_DVS}
\end{figure}

\subsubsection{Event-Based Methods}
Event cameras are a novel class of bio-inspired visual sensors~\cite{ghosh2025event}. Fig. \ref{pic_DVS} illustrates a comparison between the outputs of event cameras and those of conventional frame-based cameras. An event camera consists of an array of independently operating smart pixels, each of which continuously monitors the intensity at its own location. When a pixel detects, at time $t_k$, an intensity change relative to its previous activation time $t_{k-1}$ that exceeds a predefined threshold $C$, it asynchronously triggers an event. Let $L_{k-1}=L(x,t_{k-1})$ denote the intensity value at pixel location $x$ at time $t_{k-1}$. If $|L_{k}(x,t_k)-L_{k-1}(x,t_{k-1})|>C$, a positive-polarity event is generated; if $|L_{k}(x,t_k)-L_{k-1}(x,t_{k-1})|<-C$, a negative-polarity event is generated. Each event records the timestamp, pixel location, and polarity of an intensity change with microsecond-level temporal resolution~\cite{guo2025event}.
This results in a sparse yet high-temporal-resolution representation of dynamic scenes~\cite{wolf2025ebsnor}.

Ziegler et al. present the first real-time event-based perception pipeline for table tennis\cite{ziegler2025event}. In this pipeline, events are represented using the Exponential Reduced Ordinal Surface(EROS), and table-tennis balls are detected via Hough circle detection with \ac{ROI} constraints. Although this approach provides richer perceptual information than frame-based methods, it is limited to simplified scenes with only the ball present and lacks robustness to other moving objects. To overcome this limitation, we systematically analyze the geometric and motion characteristics of the table-tennis ball and propose a perception algorithm capable of accurately identifying the ball in complex dynamic scenes with multiple moving objects.

\subsection{Reinforcement Learning in Robotic Table Tennis} 
\ac{RL} has been widely and successfully applied to robotic table-tennis striking~\cite{huang2016jointly,gao2020robotic}. Huang et al. employed \ac{RL} to achieve accurate table tennis return control \cite{huang2016jointly}. However, their experimental training process required approximately 50,000 interaction samples. Büchler et al. adopted a model-free \ac{RL} approach combined with a pneumatically actuated robotic arm, enabling fast and safe table tennis returns, while requiring approximately 1.5 million training time steps \cite{buchler2022learning}. Similar to our work, \cite{zhu2018towards} formulated the table tennis return task as a one-step decision-making problem and introduced a Monte Carlo–based \ac{RL} method within the \ac{DDPG} framework to achieve high-precision target returns. Nevertheless, their approach still relied on approximately 200,000 interactions in simulation for training. 

The aforementioned methods all rely on a large number of environment interactions to complete policy learning, which often incurs high costs. Moreover, extended simulation usage is known to exacerbate sim-to-real gaps in highly dynamic settings. As a result, sample efficiency has emerged as a key bottleneck in \ac{RL} for robotic striking tasks. Despite this, systematic studies addressing this issue remain relatively limited.

\cite{tebbe2021sample} improved sample efficiency by having the critic network predict intermediate parameters for reward computation instead of directly estimating the reward. However, the reward function remains fixed throughout training, which leads to sparse or unstable reward gradients in the early stages of policy learning. To address this issue, we propose a human-inspired, sample-efficient strategy that combines a \ac{CDTA} reward with a reward-threshold mechanism. This approach supports progressive training from low to high ball velocities, while adaptively adjusting the reward over training episodes at each speed, thereby improving convergence and learning efficiency.

\section{METHOD}
\subsection{Problem Definition}
The table-tennis robot can be decomposed into two tightly coupled components: perception and decision-making.

The objective of the perception module is to estimate the ball’s state at the hitting plane, defined as the plane where the racket strikes the ball. This state includes the hitting time $t_c$, as well as the ball position $\mathbf{p}_c = (x_c, y_c, z_c)$ and velocity $\mathbf{v}_c = (\dot{x}_c, \dot{y}_c, \dot{z}_c)$.

The objective of the decision module is to determine the racket orientation $\boldsymbol{\theta}_c \in \mathbb{R}^3$ for hitting the ball. Following the usual practice in \ac{RL}, we define our problem as a Markov decision process (MDP) $(\mathcal{S},\mathcal{A},p,r)$. In this study, to simplify the problem, motion decision learning is modeled as a one-step decision process, with each episode consisting of a single step.
That is, the transition function $p : \mathcal{S} \times \mathcal{A} \times \mathcal{S}$ deterministically maps every state–action pair to the terminal state.
The state space \(\mathcal{S}\) is six-dimensional, consisting of the ball position and velocity: $s = (p_c, v_c) \in \mathbb{R}^6$. The action space \(\mathcal{A}\) determines the racket orientation at hitting time, which in turn controls the ball’s return direction and landing position.
$r$ is a scalar reward derived from the ball’s landing position at the end of each episode.

\begin{figure}[t]
  \centering
  \includegraphics[width=0.48\textwidth]{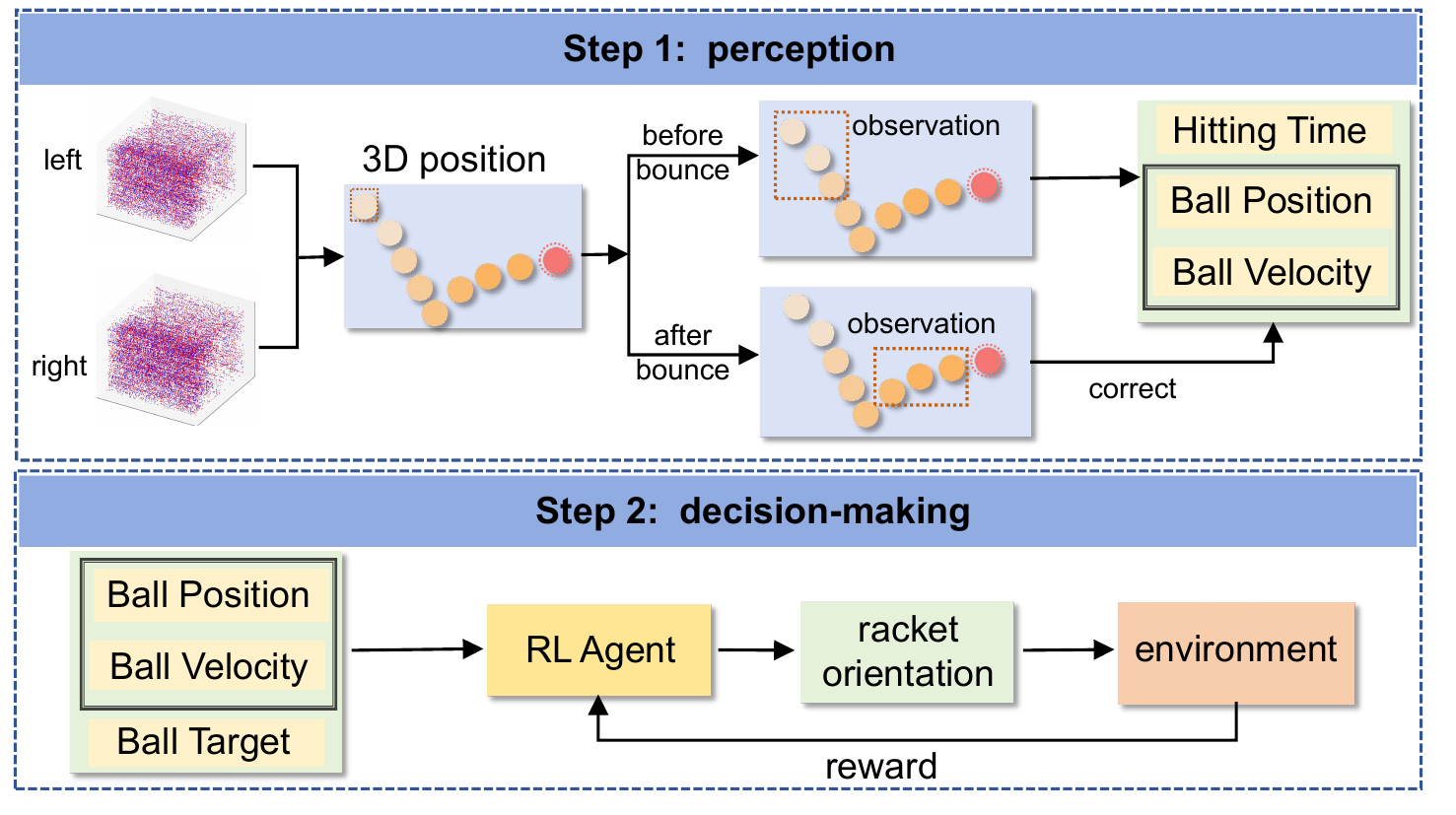}
  \caption{\textbf{Overview framework.} The simulation environment and the algorithm are connected and exchange data via the ZMQ communication protocol.}
  \label{pic_over}
\end{figure}

\begin{figure*}[t]
  \centering
  \includegraphics[width=1\textwidth]{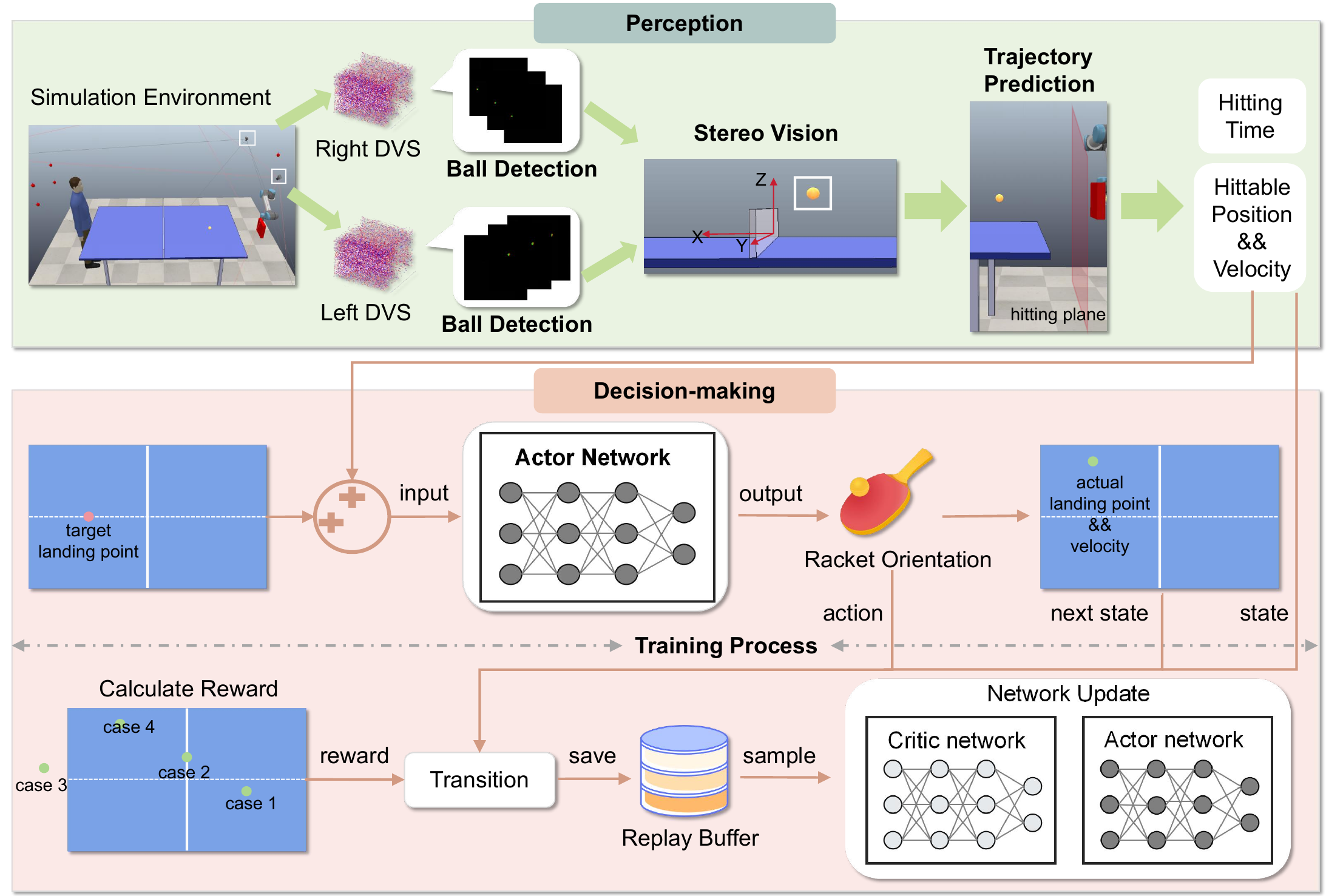}
  \caption{\textbf{Overall pipeline of the perception and decision-making modules.}}
  \label{pic_overall}
\end{figure*}

\subsection{Overall Algorithm Framework}
The overall framework is shown in Fig. \ref{pic_over}. Two event cameras are used to capture the event streams of the ball and estimate the ball’s 3D position in the world coordinate frame.

However, accurate trajectory prediction remains challenging due to the complex physical interactions during ball flight. In particular, when the ball bounces off the table surface, it experiences energy loss and abrupt changes in both velocity magnitude and direction. Since such bouncing behavior cannot be accurately modeled as an ideal elastic interaction, we adopt a two-stage trajectory prediction approach that separately models the ball motion before and after the table bounce.

In the pre-bounce stage, once five 3D position measurements have been accumulated, a polynomial fitting approach is employed to estimate the ball’s velocity and predict its future trajectory. This process yields an initial estimate of the hitting state, including the predicted hitting time $t_c$, position $\mathbf{p}_c$, and velocity $\mathbf{v}_c$. The racket is then moved toward the predicted hitting position in advance. After detecting the table bounce, a post-bounce trajectory prediction is performed using post-bounce observations, leading to a refined and more accurate estimate of the hitting state.

Once the post-bounce hitting information is obtained, the \ac{RL} agent determines the racket orientation by jointly considering the ball state and the desired target landing position. The resulting control action $\boldsymbol{\theta}_c$ is then executed in the environment to return the ball. Following the strike, a reward is computed and the agent parameters are updated accordingly. Through iterative interaction and learning, the agent converges to a policy that consistently returns the ball towards 
a desired target location.

\subsection{Perception}

The perception pipeline is illustrated in Fig.~\ref{pic_overall} 
, which consists of three main stages:
\begin{itemize}
    \renewcommand{\labelitemi}{} 
    \item \textbf{Ball Detection:} 
    Two synchronized \ac{DVS} cameras capture event streams of the ball and other scene objects. Ball detection is performed independently for each camera to obtain the 2D coordinates of the ball center.
    \item \textbf{Stereo Vision:} The 2D observations from the stereo DVS cameras are fused using stereo vision principles to estimate the 3D position of the ball in the world coordinate frame.
    \item \textbf{Trajectory Prediction:} Based on estimated 3D ball positions, the hitting time as well as the ball’s final position and velocity on the hitting plane are predicted. Trajectory prediction is executed twice during the perception process, corresponding to the pre-bounce and post-bounce phases. The same prediction method is applied in both cases.
\end{itemize}

The following sections describe the three main stages of the perception pipeline in detail.

\subsubsection{Ball Detection}

\begin{figure*}[t]
  \centering
  \includegraphics[width=1\textwidth]{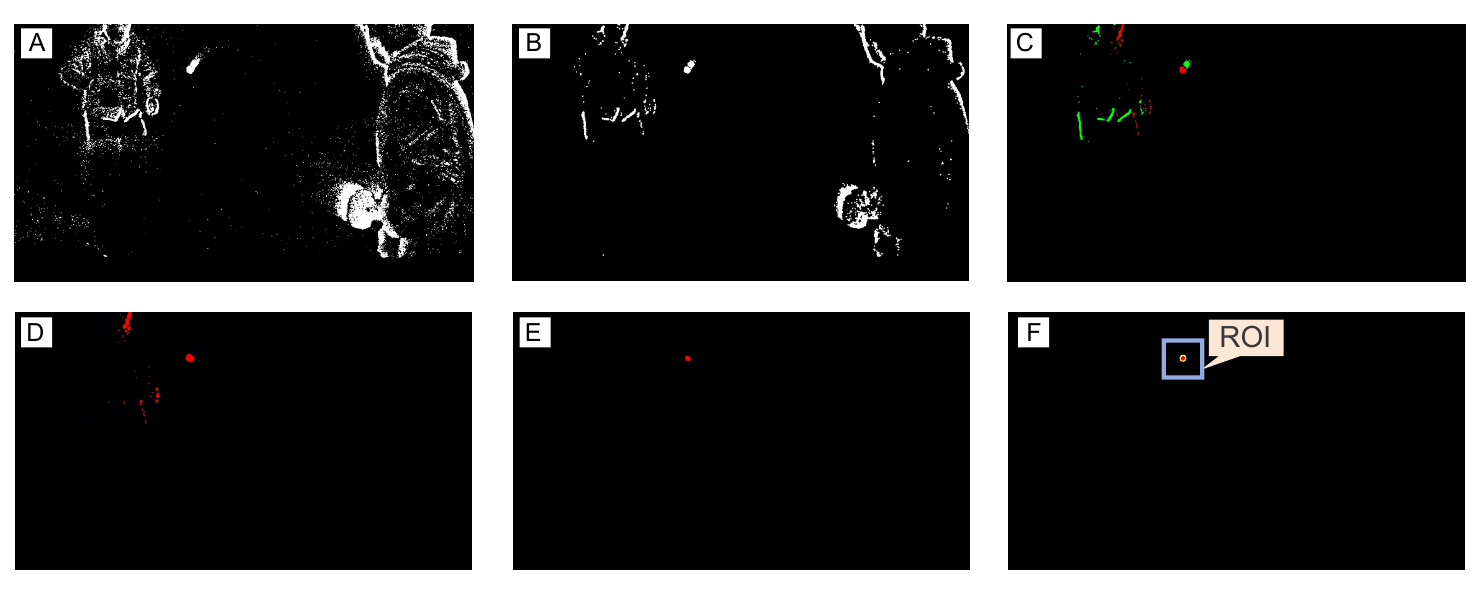}
  \caption{\textbf{Ball detection pipeline illustration.}
\textbf{(A)} Raw event image.
\textbf{(B)} Event Stream Denoising.
\textbf{(C)} Density-Based Spatial Clustering, with green indicating negative polarity and red indicating positive polarity.
\textbf{(D)} Polarity-based Event Filtering.
\textbf{(E)} Geometric Verification.
\textbf{(F)} Ball Localization and Adaptive \ac{ROI} Update.}
    \label{pic_detection}
\end{figure*}

Inspired by biological vision, the \ac{DVS} emulates the hierarchical processing of the retina by converting motion-induced contrast changes into asynchronous ON/OFF events, yielding a sparse event stream that supports rapid visual perception.
To robustly detect the ball from sparse and asynchronous event streams, we propose an event-based ball detection method that that explicitly exploits the physical characteristics of event generation.

Each event generated by the \ac{DVS} is represented as $e_i = (x_i, y_i, t_i, p_i),$
where $(x_i, y_i)$ denotes the pixel coordinates, $t_i$ is the timestamp, and $p_i \in \{0, 1\}$ indicates the polarity of the brightness change. As shown in Fig.~\ref{pic_detection}B--F and summarized in Algorithm~\ref{alg:ball_detection}, raw events are progressively refined by suppressing noise, spatiotemporal event clustering, and enforcing polarity-aware geometric constraints. To further improve detection efficiency, we define a \ac{ROI} as a local image region centered at the estimated ball position. 
At the first detection of the ball, the ROI is defined over the full sensor plane. After detecting the ball in the current iteration, we adaptively update the \ac{ROI} and restrict subsequent detections to this region.
Importantly, our algorithm operates directly on the raw event stream, reflecting the event-driven processing paradigm of the magnocellular–dorsal pathway.
This ball detection algorithm effectively isolates the ball from background events and enables reliable position estimation.

\textbf{Event Stream Denoising (see Fig. \ref{pic_detection}B):}
The raw events contain numerous isolated and spurious points (see Fig.~\ref{pic_detection}A), which primarily arise from thermal noise in the \ac{DVS} under static or low-light conditions~\cite{guo2022low}.
To remove isolated noise events and fragmented structures in event-based data, we apply a lightweight denoising strategy based on local event maps. Specifically, a local binary image is constructed within the minimum bounding region of the event set, avoiding global frame reconstruction.
We then apply a single morphological opening operation
\begin{equation}
I' = (I \ominus K) \oplus K,
\end{equation}
where $I$ denotes the binary event image and $K$ is a small elliptical structuring element.
This operation removes isolated noise while preserving compact event structures, thereby providing cleaner inputs for the subsequent clustering stage.

\textbf{Density-Based Spatial Clustering (see Fig.~\ref{pic_detection}C):}
To group spatially and temporally related events, we perform clustering on the event set. Given the event set, \ac{DBSCAN} groups points based on a neighborhood radius $\varepsilon$ and a minimum number of samples $N_{\min}$, labeling sparse points as outliers~\cite{ester1996density}. 
In addition, to incorporate a simple physical prior, we retain only clusters whose cardinality falls within 
$\left[ C_{\min},\, C_{\max} \right]$, 
corresponding to the expected number of events generated by a ball of known size moving at high speed. This criterion excludes both spurious noise clusters and overly large non-ball structures.

\textbf{Polarity-based Event Filtering (see Fig.~\ref{pic_detection}D):}
Differing from previous approaches that ignore event polarity, we exploits the polarity of events.
A key observation underlying our method is that, for a fast-moving ball, the leading edge of the event stream corresponds to the forward-facing surface of the ball and consistently generates positive-polarity events.
This occurs because the \ac{DVS} asynchronously responds to increases in log-intensity, and as the ball moves forward, its leading surface causes a local brightness increase at the corresponding pixels, triggering positive-polarity events.
Consequently, by focusing on positive-polarity events, interfering events generated during the ball’s motion are suppressed. This yields a cleaner cluster for circle fitting, ensuring accurate estimation of the ball’s center and radius even under high-speed motion.
Specifically, for each valid candidate event cluster $\mathcal{C}_k$, only positive-polarity events are retained to enhance detection accuracy, forming a refined cluster $\mathcal{C}_k^{+} = \{(x_i, y_i) \in \mathcal{C}_k \mid p_i = 1\}$.
Furthermore, we remove spatial outliers based on their Euclidean distances to the cluster centroid $(\bar{x}, \bar{y})$.
The centroid of $\mathcal{C}_k^{+}$ is computed as
\begin{equation}
(\bar{x}, \bar{y}) =
\frac{1}{|\mathcal{C}_k^{+}|}
\sum_{(x_i,y_i)\in\mathcal{C}_k^{+}} (x_i, y_i).
\end{equation}
Events whose distances exceed a predefined threshold are discarded, yielding a more compact cluster and reducing the risk of inaccurate or failed circular fitting.

\textbf{Geometric Verification (see Fig.~\ref{pic_detection}E):}
To obtain a reliable representation of the ball, we impose geometric constraints based on circularity and solidity. Specifically, for each polarity-refined cluster, we compute its convex hull and then measure the enclosed area $A_k$ and perimeter $P_k$, which are used to define the circularity metric $\gamma_k = 4\pi A_k / P_k^2$.
Clusters whose circularity $\gamma_k$ falls below a predefined threshold $\gamma_\text{min}$ are excluded.
In addition, we enforce a solidity requirement by computing $\eta_k = A_{\text{pixel}} / A_k$, where $A_{\text{pixel}}$ denotes the actual occupied pixel area of the cluster.
This constraint effectively filters out square-like or elongated structures caused by rackets or other moving objects.

\textbf{Ball Localization and Adaptive ROI Update (see Fig.~\ref{pic_detection}F):}
A minimum enclosing circle is fitted to each validated cluster to estimate the ball position. The circle yields the estimated center $(u, v)$ and radius $r$, and clusters with radii within a predefined valid range are retained as the final detection results. 

\begin{algorithm}[t]
\caption{Event-based Ball Detection Method}
\label{alg:ball_detection}
\begin{algorithmic}[1]
\Require Event stream $e_i = \{(x_i, y_i, t_i, p_i)\}$ from two DVS cameras, ROI expansion $roi\_expand$
\Ensure Estimated ball center $(u,v)$

\State Initialize ROI to full sensor plane: $x_{\min}=0, x_{\max}=W, y_{\min}=0, y_{\max}=H$

\While{new events arrive}
    \State Extract events within current ROI
    \State Event Stream Denoising
    \State Density-Based Spatial Clustering
    \State Polarity-based Event Filtering
    \State Geometric Verification
    \State Ball Localization $(u,v)$
    \If{ball detected at $(u,v)$}
        \State Update ROI around the detected ball:
        \[
            x_{\min} = u - roi\_expand, \quad x_{\max} = u + roi\_expand
        \]  
        \[
            y_{\min} = v - roi\_expand, \quad y_{\max} = v + roi\_expand
        \]
    \ElsIf{ball missed for 3 consecutive times}
        \State Reset ROI to full sensor plane
    \EndIf
\EndWhile
\end{algorithmic}
\end{algorithm}

\subsubsection{Stereo vision}
\begin{figure}[t]
  \centering
  \includegraphics[width=0.48\textwidth]{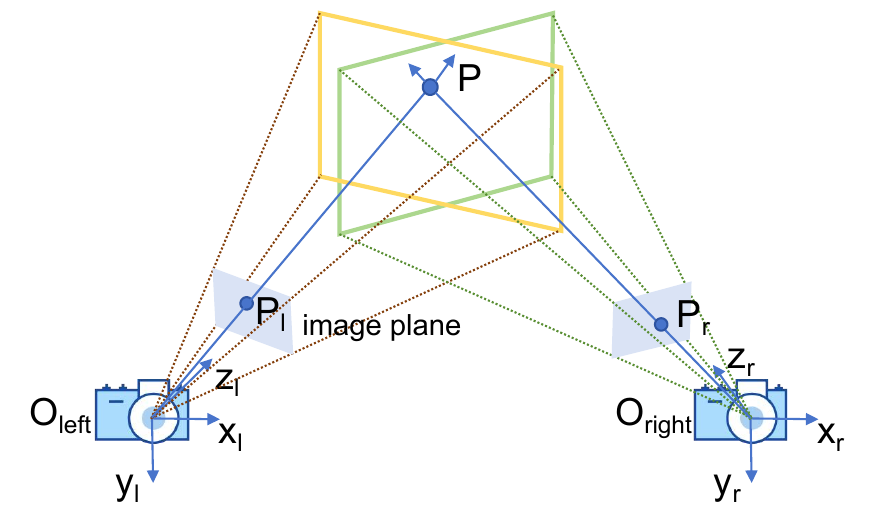}
  \caption{\textbf{Stereo vision illustration.} The system consists of two fixed event cameras whose optical center positions in the world coordinate frame are known and denoted as $O_{left}$ and $O_{right}$, respectively.}
  \label{pic_stereo}
\end{figure}

To recover the 3D position of a ball from a stereo event-camera system, we employ a triangulation method~\cite{lin2019ball}, as shown in Fig.~\ref{pic_stereo}. Initially, the detected pixel coordinates of the ball are converted into the camera coordinate frames and used to define rays originating from the optical centers of the two cameras. These rays are then transformed into the world coordinate frame using the known camera poses. Finally, the ball position is estimated as the intersection of the two rays, which is computed in a least-squares sense to account for detection noise and calibration errors.

\subsubsection{Trajectory Prediction}
To predict the ball’s state at a predefined hitting plane, we use polynomial fitting on five discrete 3D observations to estimate its future trajectory~\cite{lin1983formulation}.
Considering only the effect of gravity, the ball motion is modeled as linear in the horizontal directions and uniformly accelerated in the vertical direction. Velocity components are then estimated using least-squares fitting. Based on this model, the hitting time is computed, and the corresponding position and velocity of the ball are directly obtained. This approach leverages prior knowledge of ball dynamics, enabling accurate state estimation from sparse observations with low computational cost, suitable for real-time applications.

\subsection{Decision-making}
To improve sample efficiency in high-speed table tennis scenarios, we draw inspiration from human skill acquisition. Similar to how humans struggle to acquire skills at very high ball speeds, directly initiating policy learning under excessively fast conditions often makes the training process difficult to converge. Therefore, we first train the policy in low-speed scenarios and adopt a gradual skill acquisition approach, in which the agent initially learns to return the ball onto the table and subsequently learns to return it toward a target location. Once convergence is achieved at lower speeds, the policy is further trained under higher-speed conditions. This progressive training scheme enables faster and more stable convergence with significantly improved sample efficiency. In this process, We introduce two novel components, a \ac{CDTA} reward and a reward-threshold mechanism, designed to enhance learning efficiency. In the following sections, we detail the design and implementation of these components, as well as our \ac{RL} setup, including exploration strategy, actor, and critic networks.

\subsubsection{Case-dependent Temporally-adaptive Reward}
\begin{figure}[t]
  \centering
  \includegraphics[width=0.48\textwidth]{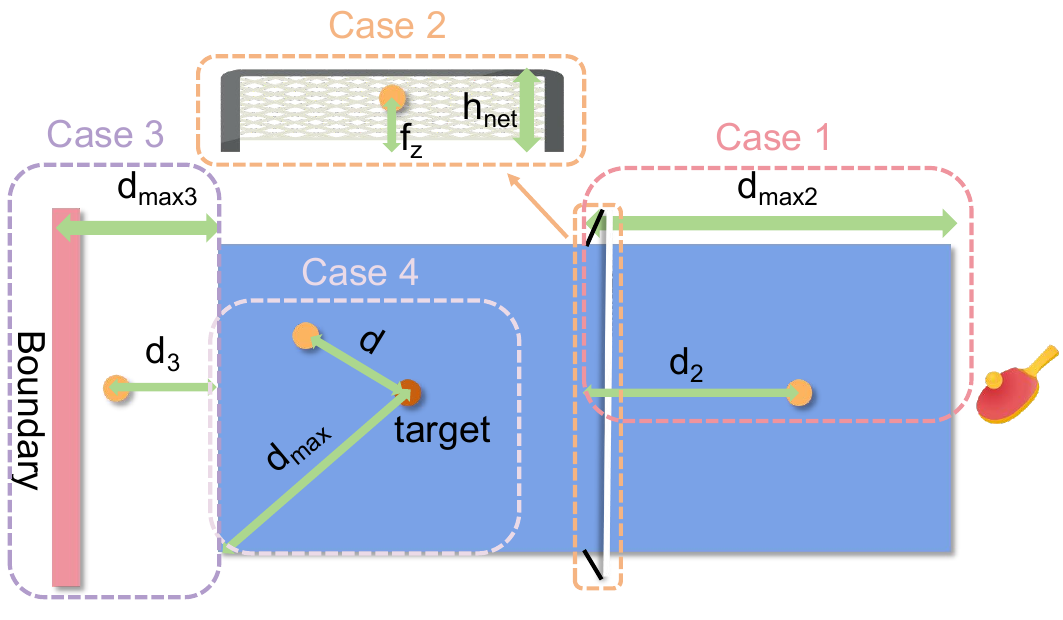}
  \caption{\textbf{Reward design illustration.} The figure shows the reward assignments corresponding to four different ball landing cases. The boundary indicated in the figure is an artificially defined valid region; once the ball lands outside this boundary, the simulation is reset and a new serve is initiated.}
  \label{pic_reward}
\end{figure}

To promote skill progression at each ball velocity and improve sample efficiency, we introduce a \textbf{\ac{CDTA} reward} that gradually shifts from basic to advanced skills. The training objective initially focuses on successful table returns, then gradually shifts to emphasize accuracy toward the target landing location. 
Accordingly, we propose a novel reward design to realize the proposed \ac{CDTA} reward.
Fig.~\ref{pic_reward} provides an intuitive illustration of our reward design, highlighting the four distinct ball landing cases. Cases~1–3 correspond to failed returns, including landing on the player’s side of the table, hitting the net, or going off the table, while Case~4 corresponds to a successful return.
The overall reward formulation is defined as follows:
\begin{equation}
R = \begin{cases} 
- \lambda_1 \cdot \dfrac{d_2}{d_{max2}} 
& \text{Case1}, \\[6pt]

- \lambda_2 \cdot \left( 1 - \dfrac{f_z}{h_{net}} \right) 
& \text{Case2}, \\[6pt]

- \lambda_3 \cdot \dfrac{d_3}{d_{max3}} 
& \text{Case3}, \\[6pt]

\sum_{i=1}^{2} w_i(n) \cdot r_i(f_x, f_y, f_z) \quad 
& \text{Case4}.
\end{cases}
\end{equation}
where the reward is computed based on the ball landing outcome observed within each episode (Fig.~\ref{pic_overall}). 
$\lambda_i > 0$ $(i=1,2,3)$ are predefined penalty coefficients.

For Case~4, we employ a weighted combination of two landing-based reward terms.
The episode-dependent weights enable an automatic transition of the reward emphasis as training progresses. The weights \(w_i(n)\) are functions of the training episode \(n\) and are given by
\begin{equation}
w_i(n) =
\begin{cases}
e^{-\beta n}, & i = 1, \\[8pt]
1 - e^{-\beta n}, & i = 2.
\end{cases}
\end{equation}
where \(\beta > 0\) is a hyperparameter controlling the exponential decay rate. The landing-based reward term \(r_i(\cdot)\) is computed based on the landing error as
\begin{equation}
r_i(\cdot) =
\begin{cases}
\lambda_{\text{base}} , 
& i = 1, \\[8pt]

\lambda_{\text{base}} + \lambda_{\text{scale}} \cdot 
\max \left( 0, 1 - \dfrac{d}{d_{\max}} \right), 
& i = 2.
\end{cases}
\end{equation}
where \(\lambda_{\text{base}}\) and \(\lambda_{\text{scale}}\) are reward coefficients, and \(d\) and \(d_{\max}\) denote the distance measures illustrated in Fig.~\ref{pic_reward}.

The proposed reward design enables a smooth and continuous adaptation of the reward structure over the course of training. 
In the initial phase of training, the weight $w_1(n)$ dominates, emphasizing the base reward for achieving successful table returns and encouraging the agent to learn basic return capability.
With increasing training episodes $n$, $w_1(n)$ decreases, diminishing the weight of $r_1$ for successful table returns.  
Meanwhile, $w_2(n)$ increases, progressively highlighting $r_2$ to enable accurate returns to the target point.
In this manner, the reward structure evolves progressively under a unified formulation, while avoiding discontinuities or explicit stage transitions.

Moreover, in the \ac{CDTA} reward, it is crucial to incorporate case-based rewards to guide learning from failed return cases. Particularly during the initial phase of training, the policy remains highly stochastic, resulting in frequent failed return attempts. To provide more informative learning signals, we employ case-based rewards formulation, which explicitly accounts for different return landing cases. These rewards encourage returns closer to the target region on the opponent's side, providing clear guidance for policy improvement.

\subsubsection{Reward-threshold Mechanism}
After converging at low speeds, the policy continues training in higher-speed scenarios. However, continuing training from an already converged policy introduces a critical challenge related to the effective utilization of experience accumulated in prior training stages. In response, we explored two intuitive strategies:
\begin{itemize}
    \renewcommand{\labelitemi}{} 
    \item \textbf{Buffer discarding}:  Discards all transitions stored in the replay buffer.
    \item \textbf{Buffer retention}: Keeps all past transitions in the replay buffer.
\end{itemize}

Each of the two straightforward strategies for handling the replay buffer has its limitations. Discarding the entire replay buffer results in the loss of valuable successful transitions. In contrast, retaining all past transitions preserves these successful transitions while also including failed transitions from prior training, which may interfere with current policy optimization. Consequently, neither strategy alone provides an ideal balance between leveraging past knowledge and avoiding detrimental transitions. 

To address this limitation, we propose a \textbf{reward-threshold mechanism}, which retains only high-quality transitions. As illustrated in Fig.~\ref{pic_overall}, each transition stored in the replay buffer is defined as $\tau_i = (s_i, a_i, r_i, s_i'),$
where $s_i$, $a_i$, $r_i$, and $s_i'$ denote the state, action, reward, and next state, respectively. Specifically, the next state $s_i'$ corresponds to the ball’s actual landing position and velocity. For all transitions collected in previous training stages, we introduce a reward threshold $\delta$ to characterize high-quality transitions (i.e., the ball lands on the opponent’s table). This implies that only transitions with rewards above $\delta$ are retained in the replay buffer. As a result, we consider two sources of transitions for the replay buffer during continued training at higher ball speeds: 
(i) high-reward transitions inherited from the previous model ($r_i \ge \delta$), and 
(ii) new transitions collected under the current training conditions, that is,
\begin{equation}
\mathcal{D}_{\text{train}}
=
\mathcal{D}_{\text{new}} \cup
\left\{
\tau_i \in \mathcal{D}_{\text{old}} \mid r_i \ge \delta
\right\}.
\end{equation}

Using this combined buffer, the Critic network is trained by minimizing the standard TD loss:
\begin{equation}
\mathcal{L}(\theta^Q)
=
\mathbb{E}_{(s,a,r,s') \sim \mathcal{D}_{\text{train}}}
\Big[
\big(
Q(s,a,g;\theta^Q) - y
\big)^2
\Big],
\end{equation}
where the target value $y$ is computed as
\begin{equation}
y = r + \gamma Q'\!\left(s', \mu'(s')\right).
\end{equation}

Correspondingly, the Actor network is optimized to maximize the expected action value estimated by the Critic:
\begin{equation}
J(\theta^\mu)
=
\mathbb{E}_{s \sim \mathcal{D}_{\text{train}}}
\left[
Q\big(s,g, \mu(s;\theta^\mu)\big)
\right].
\end{equation}

This approach has two key advantages. First, it avoids the negative impact of failed experiences while retaining valuable successful ones, thereby focusing learning on effective transitions. Second,  it provides a generalizable strategy that can be applied to other \ac{RL} tasks.

\subsubsection{Exploration}
We employ an exponentially decaying exploration strategy to balance exploration and exploitation during training. During training, actions are perturbed with additive Gaussian noise that decays progressively, defined as
\begin{equation}
\eta = \max(\eta_{\min}, \eta_0 \cdot \gamma^{\text{n}}),
\end{equation}
where 
$\eta$ denotes the current exploration noise magnitude, 
$\eta_0$ is the initial noise magnitude, 
$\eta_{\min}$ denotes the minimum permissible noise, 
$\gamma < 1$ is the decay rate, and 
$n$ denotes the training episode.

This formulation enables a gradual transition from exploration to exploitation.  
Initially, larger noise facilitates broad exploration, helping the agent discover effective strategies.
As the training episode $n$ increases, the noise is progressively reduced, allowing the agent to exploit the learned policy more reliably. 
To account for the physical limitations of the racket's workspace, we set the initial exploration noise to 0.1, ensuring that early actions remain within feasible operational bounds.

\subsubsection{Actor}
The actor network is implemented as a two-layer \ac{MLP}, which outputs deterministic racket orientations $\mu(s, g)$ (Fig.~\ref{pic_overall}). 
The network takes as input the concatenated vector of the ball state $s$ and the target landing position $g$.
Following processing by the hidden layers, the output layer employs a \texttt{tanh} activation to bound the action within $[-1,1]$.
The normalized output is then scaled by a maximum allowed rotation angle of $15^\circ$ to produce the incremental Euler angles for racket control, following an intrinsic $x$--$y$--$z$ rotation order.
To obtain the final quaternion action, we combine these incremental Euler angles with a base quaternion representing the initial racket orientation, producing a smooth and physically feasible racket motion.

\subsubsection{Critic}
The Critic network is also implemented as a two-layer \ac{MLP}, which estimates the action-value function 
$Q(s, g, a)$. 
The network takes as input the concatenated vector of the ball state $s$, target landing position $g$, and action $a$, 
and outputs a scalar representing the estimated Q-value.

\section{EXPERIMENTS}

\begin{figure*}[t]
  \centering
  \includegraphics[width=1\textwidth]{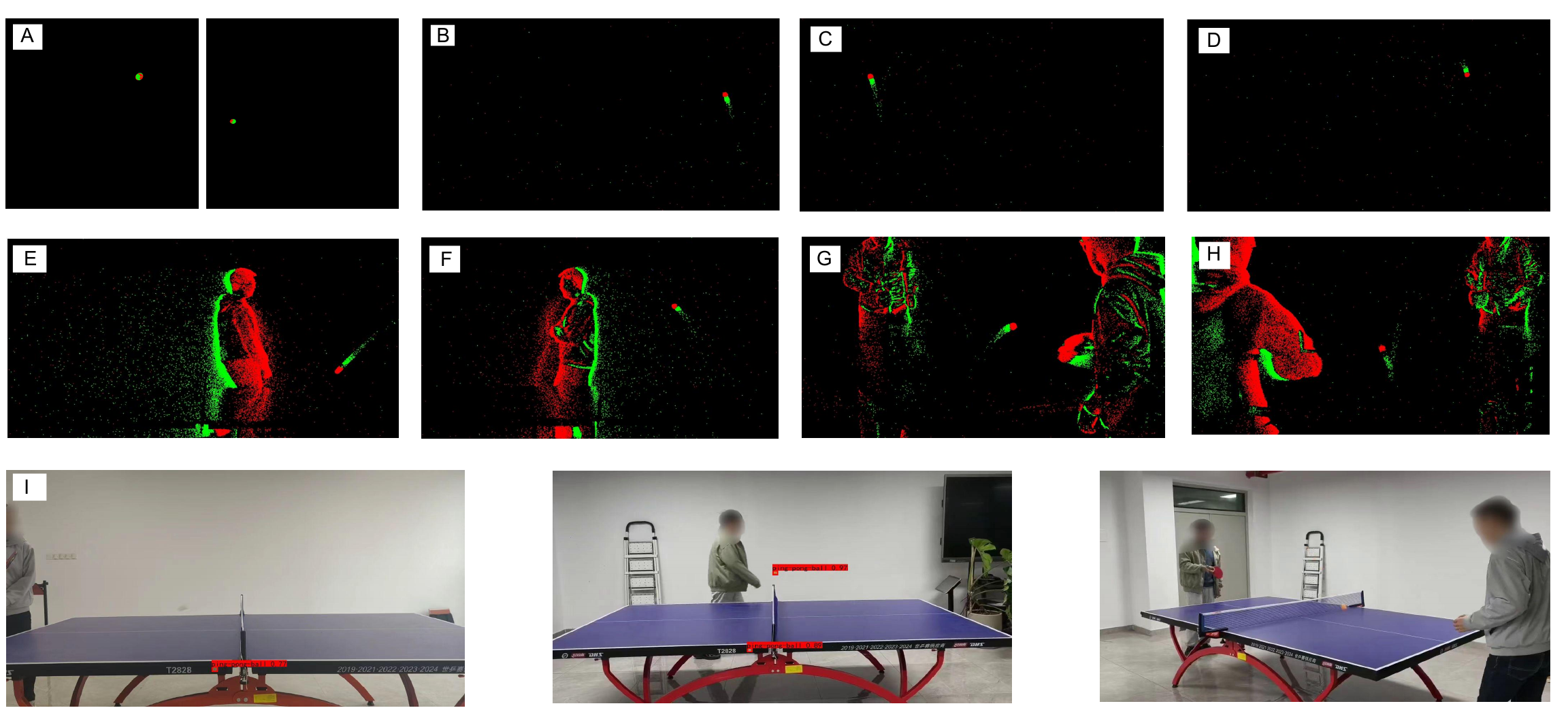}
  \caption{\textbf{Dataset illustration.}
\textbf{(A)} Imaging examples of the left and right DVS cameras in the simulated dataset.
\textbf{(B--H)} Examples from the real-world dataset, where \textbf{B}, \textbf{C}, and \textbf{D} correspond to scenes containing only the table tennis ball, and the remaining scenes include interference from moving humans.
\textbf{(I)} Examples from the RGB dataset, illustrating that RGB-based detection is more susceptible to interference caused by motion blur and static circular objects in the background.
}
  \label{pic_data}
\end{figure*}
\subsection{Data Collection and Processing}

The limited availability of event-based perception datasets for table tennis has become a key bottleneck hindering the development of event-based perception and learning methods. To address this limitation, we perform data collection across both simulated and real-world settings. In the high-fidelity simulation environment, we collect DVS event streams under a range of ball speed settings, including 5, 6, 7, 8, and 9 m/s. In addition, the ball’s ground-truth position in the world coordinate frame is recorded in real time. In real-world scenarios, we use two DVSync cameras with a resolution of 1280×720 to capture event streams of the ball motion. To thoroughly assess the feasibility of the proposed method, we collected data across seven distinct settings, covering different camera placements, diverse human motion trajectories, and actual rally play, thereby constructing a high-quality dataset. In addition to DVS data, RGB data are collected under the same experimental scenarios using an RGB camera with 
a resolution of 1280 × 720, consistent with the DVS sensor. This enables comparative studies between RGB and DVS methods. However, the RGB camera in the simulation environment does not model real camera exposure, resulting in imaging properties that differ from those observed in real-world settings. Consequently, we do not collect RGB data in simulation. An overview of the collected dataset in both simulation and real-world environments is illustrated in Fig.~\ref{pic_data}.

Due to the lack of reliable access to ground-truth ball positions in the world coordinate frame in real-world settings, we adopt manually annotated 2D ball locations in the pixel coordinate frame as ground truth. Specifically, the ball positions are labeled using LabelImg.
For the DVS data, a frame reconstruction method is applied to convert event streams into frame-based representations, on which the same annotation process is performed. In total, the dataset includes DVS and RGB data collected in real-world scenarios, comprising 2,093 ground-truth 2D ball coordinates from DVS data and 364 ground-truth 2D ball coordinates from RGB data, as well as DVS data collected in the simulation environment with 2,789 ground-truth 3D ball coordinates.
We have made the dataset publicly available at: \url{https://drive.google.com/drive}

\subsection{Robot Table Tennis Simulation Setup}

\begin{figure}[t]
  \centering
  \includegraphics[width=0.48\textwidth]{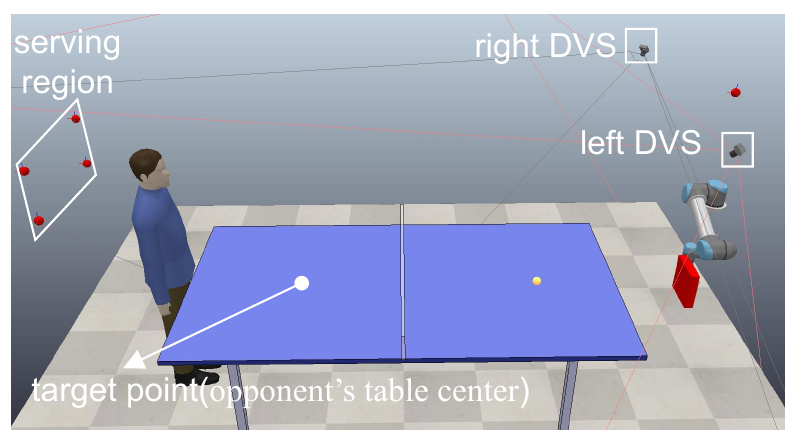}
  \caption{\textbf{High-fidelity table-tennis robot in a simulation environment.} }
  \label{pic_man}
\end{figure}

We develop a high-fidelity table tennis simulation environment on the CoppeliaSim platform (see Fig.~\ref{pic_man}). In the simulation, we defined the ball’s initial position range and initial velocity, and considered only gravity during the ball’s motion. Collisions between the ball and the table or paddle were not modeled as simple perfectly elastic collisions, but rather included friction to better approximate real interactions. To mitigate penetration issues commonly encountered in simulation, we increase the geometric thickness of the paddle to ensure stable collision detection. Additionally, the environment is bounded, and the ball is reinitialized and served again when it crosses the boundaries.

For perception, two DVS cameras with a resolution of 1280×1280 are employed, and the simulation dt is set to 0.001s to approximate the temporal characteristics of real DVS cameras. While this high temporal resolution yields fine-grained perception data, it substantially increases the computational burden of the simulation. As a result, the ball is unable to reach realistic velocities, which in turn adversely affects policy training.
To address this issue during decision-making training, the simulation dt is increased to 0.01s. In this setting, we directly obtain only the ball’s ground-truth 3D position from the simulation, while the perception algorithm described above predicts the subsequent trajectory to determine the hitting point.

\section{RESULTS}
\subsection{Accurate and Low-Latency Perception}

\begin{figure*}[t]
    \centering

    \vspace{-0.2cm}
    \begin{tikzpicture}
        \matrix[
            column sep=3mm,
            row sep=1mm
        ] {
            \node[minimum height=0.12cm, minimum width=0.7cm, fill=RGBColor] {}; &
            \node {\scriptsize RGB}; &
            \node {}; &
            \node[minimum height=0.12cm, minimum width=0.7cm, fill=DVSColor] {}; &
            \node {\scriptsize DVS(Ours)}; &
            \node {}; \\
        };
    \end{tikzpicture}
    \vspace{-0.25cm}

    \subfloat[\scriptsize Size $\downarrow$]{
        \includegraphics[width=0.24\linewidth]{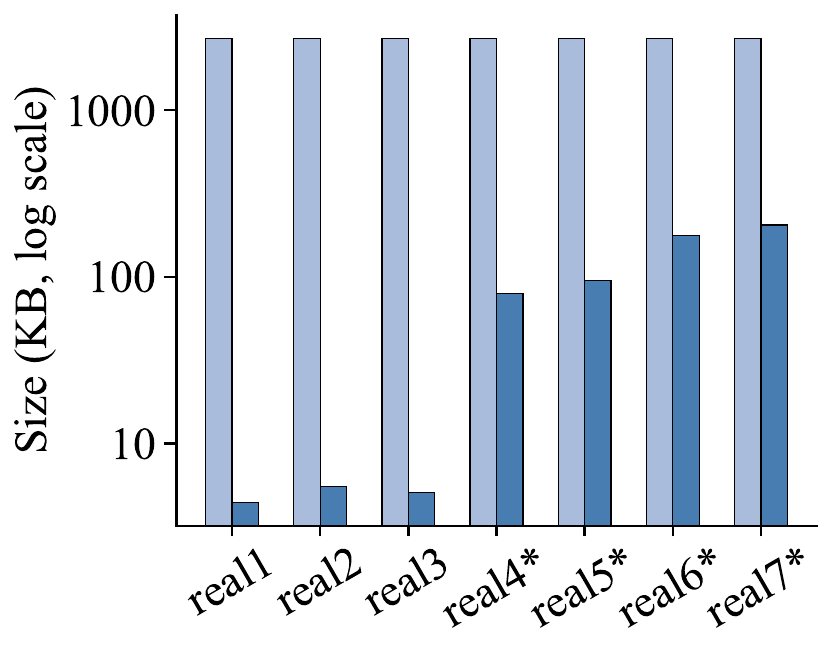}
    }
    \hfill
    \subfloat[\scriptsize Latency $\downarrow$]{
        \includegraphics[width=0.24\linewidth]{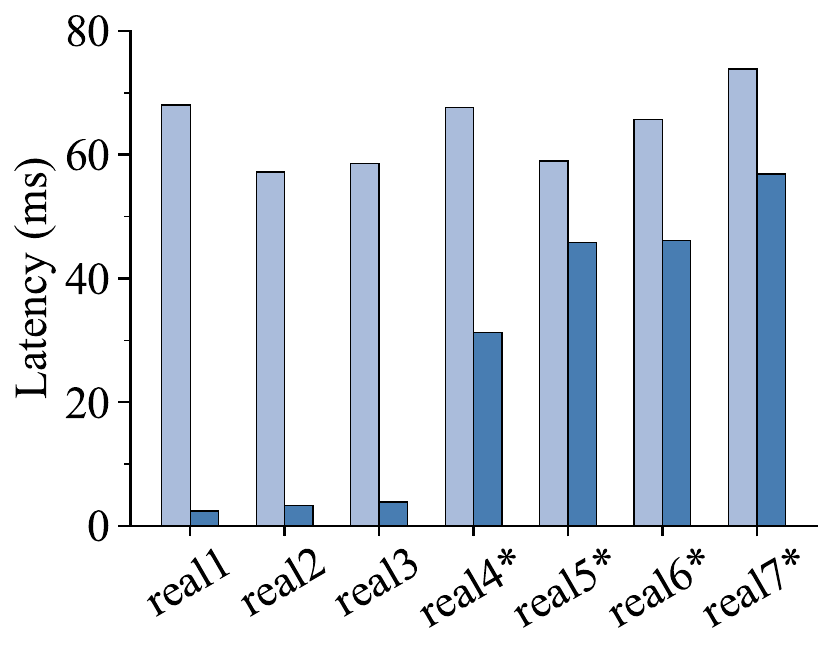}
    }
    \hfill
    \subfloat[\scriptsize Precision $\uparrow$]{
        \includegraphics[width=0.24\linewidth]{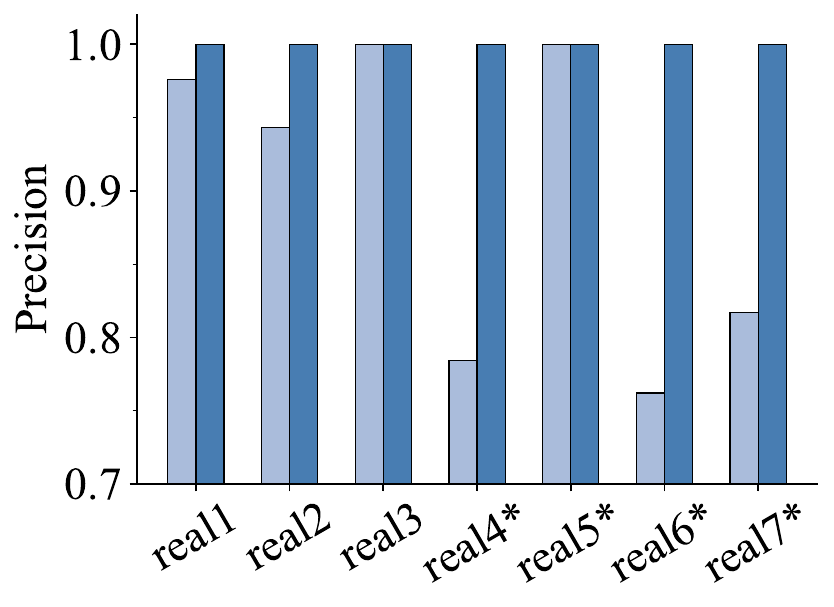}
    }
    \hfill
    \subfloat[\scriptsize Recall $\uparrow$]{
        \includegraphics[width=0.24\linewidth]{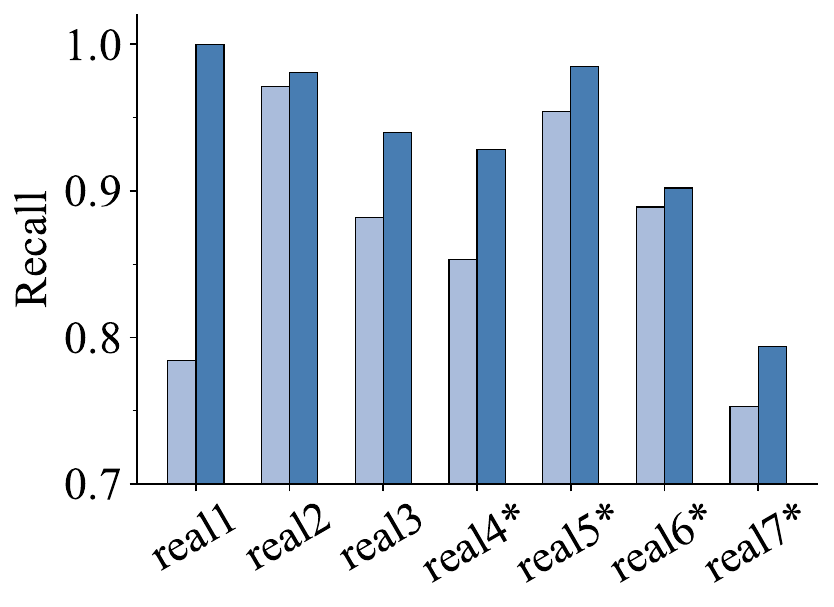}
    }

    \caption{\textbf{Comparison of detection performance between RGB and DVS across different scenes.} Datasets marked with an asterisk (*) contain multiple moving objects, whereas datasets without an asterisk include only the ball.}
    \label{fig:rgb_dvs_bars}
\end{figure*}
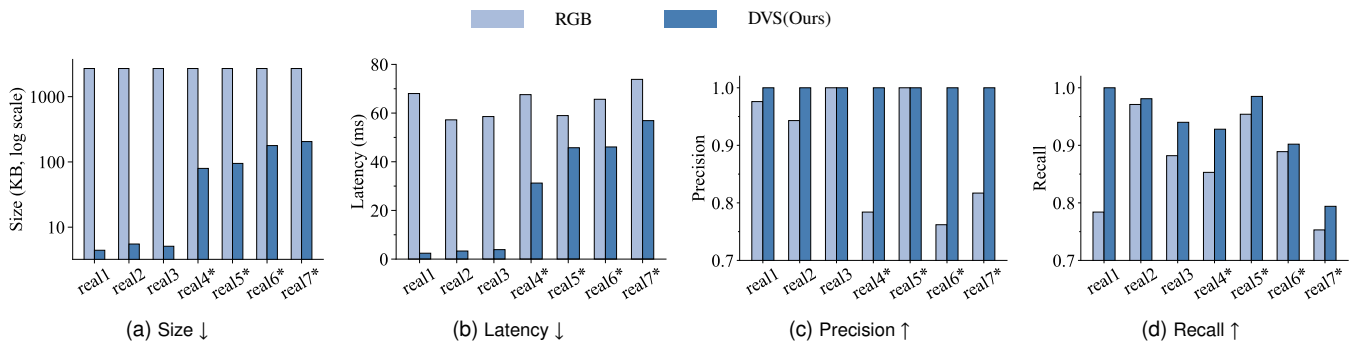

\begin{figure*}[t]
    \centering

    \vspace{-0.2cm}
    \begin{tikzpicture}
        \matrix[
            column sep=3mm,
            row sep=1mm
        ] {
            \node[minimum height=0.12cm, minimum width=0.7cm, fill=fusionColor] {}; &
            \node {\scriptsize F-E Fusion}; &
            \node {}; &
            \node[minimum height=0.12cm, minimum width=0.7cm, fill=EBPPColor] {}; &
            \node {\scriptsize EBPP}; &
            \node {}; &
            \node[minimum height=0.12cm, minimum width=0.7cm, fill=oursColor] {}; &
            \node {\scriptsize Ours}; \\
        };
    \end{tikzpicture}
    \vspace{-0.25cm}

    \subfloat[\scriptsize Precision $\uparrow$]{
        \includegraphics[width=0.24\linewidth]{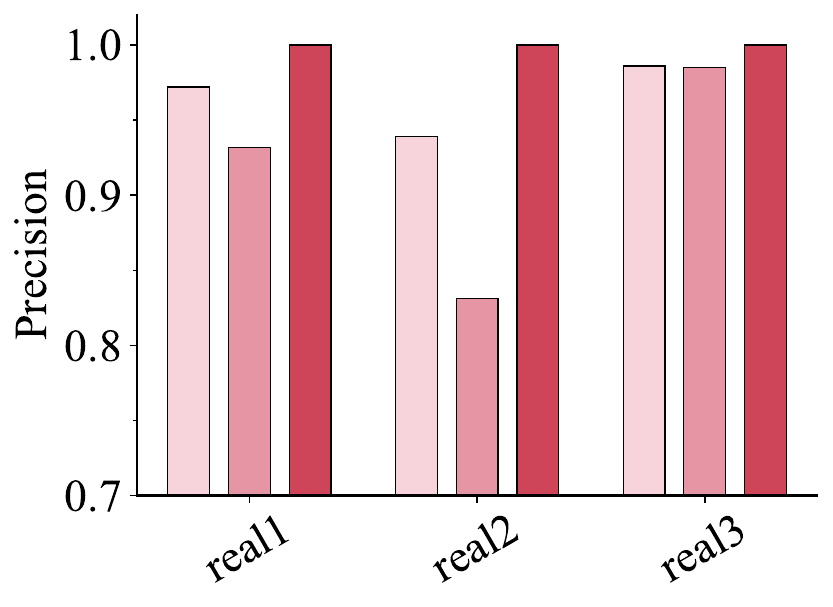}
    }
    \hspace{0.02\textwidth}%
    \subfloat[\scriptsize Recall $\uparrow$]{
        \includegraphics[width=0.24\linewidth]{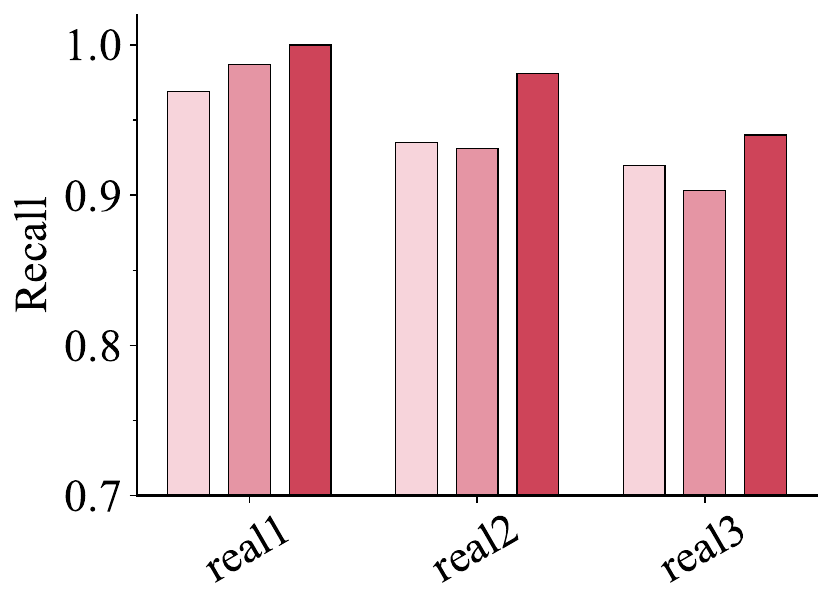}
    }
    \hspace{0.02\textwidth}%
    \subfloat[\scriptsize Error (px) $\downarrow$]{
        \includegraphics[width=0.24\linewidth]{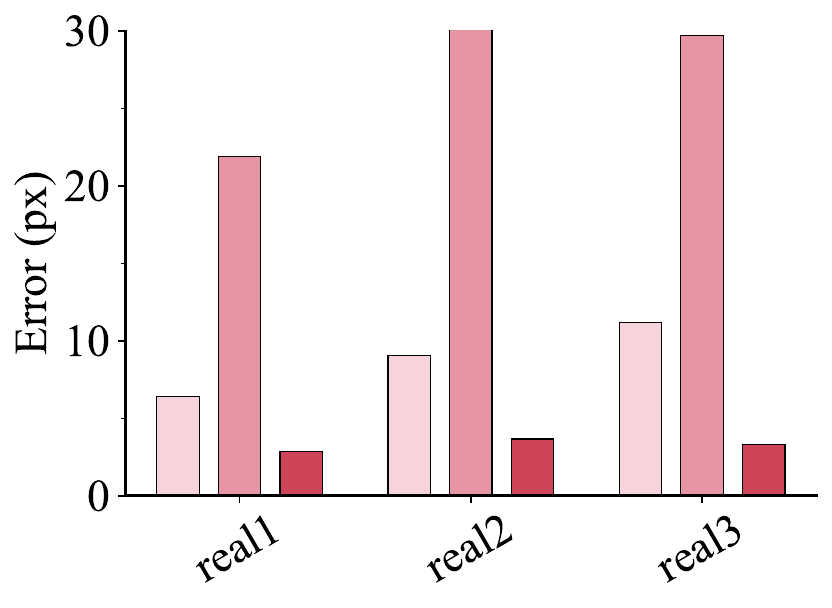}
    }

    \caption{\textbf{Comparison of different event-based methods on real-world datasets.} As F-E Fusion and EBPP operate only in scenes where the ball is present, the comparison is limited to the first three ball-only scenes.}
    \label{fig:dvs_real_bars}
\end{figure*}
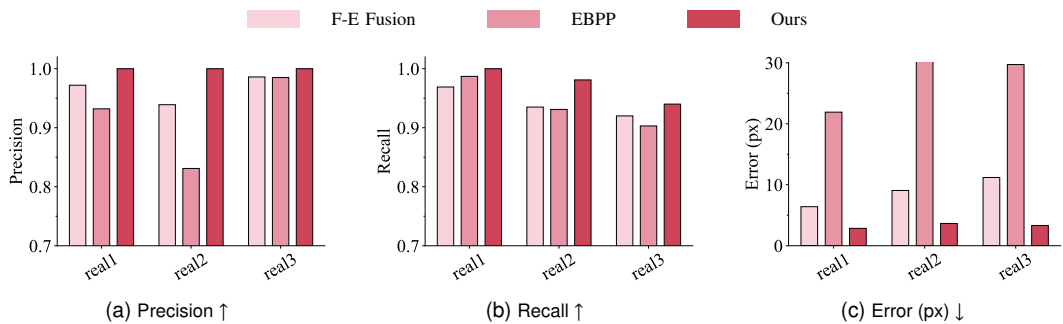

In the following experiments, we compare the proposed perception pipeline with three baseline methods, with all comparisons conducted on real-world datasets.  
The first is a frame-based approach, in which Wang et al. employ \textbf{YOLOv4-tiny} to detect balls in RGB images \cite{wang2025spikepingpong}. The other two are event-based methods. One baseline is the approach proposed by Xiao et al. \cite{xiao2023fe}, which performs DVS object detection using \ac{DBSCAN} clustering (referred to as \textbf{F-E Fusion}). In addition to \ac{DBSCAN}, the original method employs a dynamic neural field (DNF) to suppress thermal noise and emphasize moving objects. However, due to the large number of neurons involved, the DNF incurs substantial computational overhead and results in low processing speed, making it unsuitable for real-time ball detection. Therefore, in our implementation, we optimize this baseline by removing the DNF. After clustering, the ball center is estimated as the centroid of each event cluster, and the ball radius is defined as the average distance from the events to the cluster center. 
Another baseline is is the method proposed by Ziegler et al., which combines EROS with Hough circle detection for ball detection (referred to as \textbf{EBPP}) \cite{ziegler2025event}. 

To ensure a fair comparison, the parameters of all methods were carefully tuned. For latency evaluation, all methods were implemented in Python and evaluated on the same laptop equipped with an AMD Ryzen 7 8845H CPU, ensuring consistent experimental conditions.

\subsubsection{Comparison with Frame-based Methods}
Our method achieves higher perception accuracy with lower latency and reduced data processing compared to Frame-based approaches by adopting a DVS-based perception pipeline, as shown in Fig.~\ref{fig:rgb_dvs_bars}. Frame-based methods require processing the entire image at each detection step, whereas event-based methods process only sparse event streams generated by intensity changes. Consequently, our method reduces the data size required for each ball detection by up to 99.8\% relative to RGB-based methods.
Furthermore, YOLOv4-tiny is a neural-network-based method that incurs relatively high computational overhead during inference. 
In contrast, our method achieves lower latency, reducing it by up to 96.4\% compared with the RGB baseline and demonstrating superior real-time performance.

With respect to detection accuracy, the RGB method exhibits lower precision and recall, indicating a higher incidence of false positives and missed detections. This primarily results from static circular objects in the background and severe motion blur of the ball under high-speed motion, both of which can lead to detection errors, as illustrated in Fig.~\ref{pic_data}I.

\subsubsection{Comparison with Event-based Methods}
Our method achieves superior perception accuracy and robustness over existing event-based \ac{SOTA} methods in real-world scenarios by leveraging motion cues and geometric features, as summarized in Fig.~\ref{fig:dvs_real_bars}. Since the baseline methods can only operate in scenarios containing the ball alone, comparisons in the figure are restricted to such cases. The experimental results indicate that our method outperforms the other two approaches in terms of precision, recall, and localization error. The improvement is primarily due to the substantial background noise, which can interfere with clustering and Hough circle detection, whereas our method effectively suppresses such noise through morphological operations and multiple geometric constraints.

To evaluate the effectiveness of our algorithm in complex dynamic scenes, Table~\ref{dvsours} reports only the results of our method on four such challenging scenarios. Overall, the performance remains comparable to that achieved in simple ball-only scenes, demonstrating strong robustness to background dynamics.

\begin{table}[t]
\centering
\renewcommand{\arraystretch}{1.15}
\setlength{\tabcolsep}{6pt}
\caption{\textbf{Performance of our method in real-world scenes with multiple moving objects.}}
\label{tab:dvs_real}

\begin{tabular}{c|c|c|c}
\toprule
\textbf{Dataset} 
& \textbf{Precision}
& \textbf{Recall}
& \textbf{Error (px)}\\
\midrule
\multirow{1}{*}{real4*}
& \textbf{1.000} & \textbf{0.928} & \textbf{2.61} \\
\midrule

\multirow{1}{*}{real5*}
& \textbf{1.000} & \textbf{0.985} & \textbf{2.77} \\
\midrule

\multirow{1}{*}{real6*}
 & \textbf{1.000} & \textbf{0.902} & \textbf{2.59} \\
\midrule

\multirow{1}{*}{real7*}
& \textbf{1.000} & \textbf{0.794} & \textbf{2.89} \\

\bottomrule
\end{tabular}
\label{dvsours}
\end{table}

\subsection{Sample-Efficient Policy Learning}
\subsubsection{Overall Performance}
We improve sample efficiency from two aspects—the \ac{CDTA} reward and the reward-threshold mechanism—to enable a more gradual and progressive learning process. To evaluate the sample efficiency of the proposed \ac{RL} framework, we conduct comparative experiments in a high-speed scenario. 
In this setting, the ball velocity is fixed at 5 m/s, which is higher than the velocities considered in most existing works (typically in the range of 3–4 m/s), while the initial ball positions are randomly sampled for each episode~\cite{tebbe2021sample,cursi2024safe,huang2011trajectory}. 
We compare our method with several \ac{SOTA} algorithms~\cite{schulman2017proximal,babaeizadeh2016reinforcement,schulman2015trust,fujimoto2018addressing,haarnoja2018soft}. All methods are trained for the same number of episodes to ensure a fair comparison. For our method, the policy is first trained for 800 episodes at 3 m/s, and then further optimized for 900 episodes at 5 m/s. During training at each speed, we employ the proposed \ac{CDTA} reward to provide appropriate learning guidance.
Before transitioning to the 5 m/s scenario, we apply a reward-threshold mechanism to facilitate adaptation to high-speed training by retaining only experiences with successful table returns.

\begin{figure}[t]
  \centering
  \includegraphics[width=0.48\textwidth]{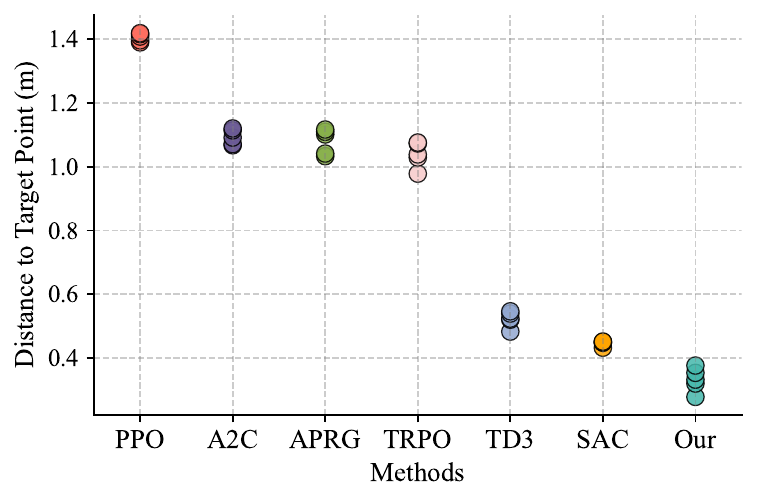}
  \caption{\textbf{Comparison against Baseline Algorithms in the High-speed Setting (5 m/s).}}
  \label{all_performance}
\end{figure}

The experimental results on return accuracy are shown in Fig.~\ref{all_performance}.
We perform evaluations every 50 episodes during the 1,700 training episodes, and each method is represented by the five best-performing evaluation results. The vertical axis denotes the distance between the returned ball and the target point. After a total of 1700 training episodes, the proposed method achieves a final average return-to-target error of 277 mm, outperforming all baseline methods by at least 35.8\%. These results clearly indicate that the proposed approach substantially improves sample efficiency in high-speed table tennis scenarios. Additional results are provided in the supplementary video.

\subsubsection{Facilitating Progressive Skill Acquisition with Case-dependent Temporally-adaptive Rewards}
\begin{figure}[t]
  \centering
  \includegraphics[width=0.48\textwidth]{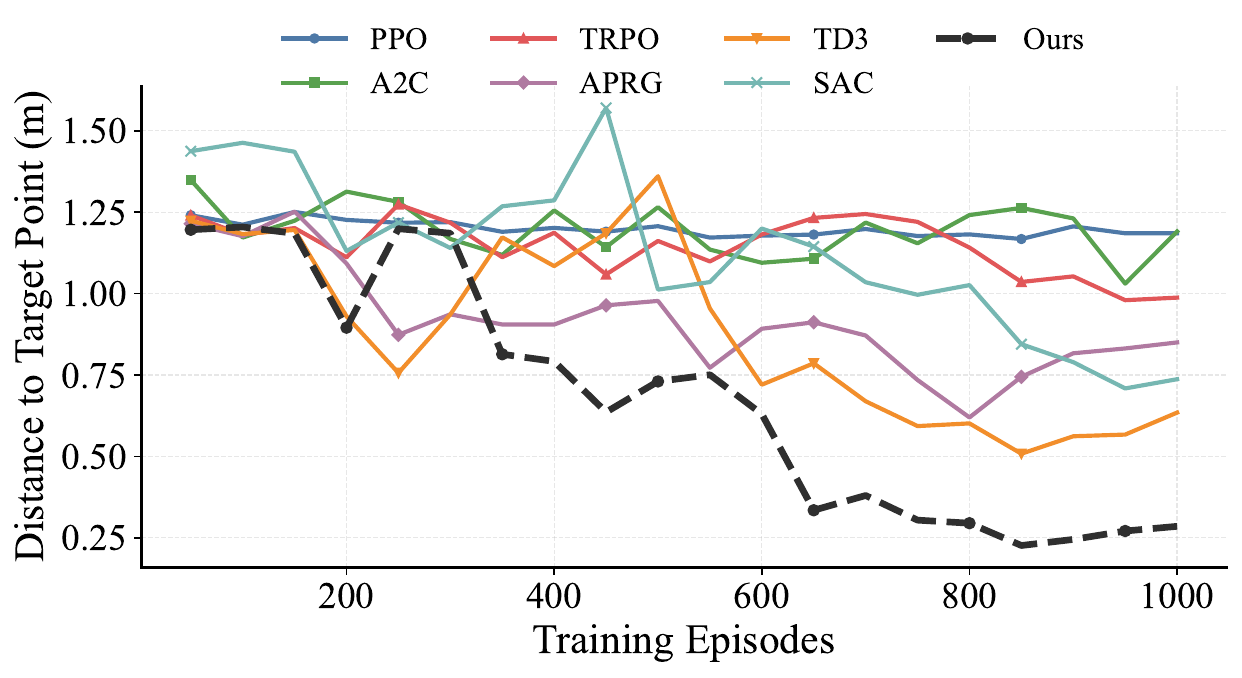}
  \caption{\textbf{Performance comparison against baseline algorithms over 1000 training iterations at 4 m/s.}}
  \label{pic_all}
\end{figure}

\begin{table}[t]
\centering
\caption{\textbf{Successful return rates (\%) under different reward designs.}}
\renewcommand{\arraystretch}{1.25}
\setlength{\tabcolsep}{6pt}
\begin{tabular}{llccc}
\toprule
\textbf{Setting} & \textbf{Episodes} & \textbf{Simple} & \textbf{Case-based Reward} & \textbf{Ours} \\
\midrule
\multirow{3}{*}{Rand pos., 3 m/s} 
 & 350  & 46\% & 70\% & \textbf{78\%} \\
 & 400  & 86\% & 80\% & \textbf{98\%} \\
 & 600 & 92\% & 98\% & \textbf{100\%} \\
\midrule
\multirow{3}{*}{Rand pos., 4 m/s} 
 & 600  & 46\% & 60\% & \textbf{72\%} \\
 & 800  & 60\% & 80\% & \textbf{92\%} \\
 & 900 & 60\% & 82\% & \textbf{98\%} \\
\bottomrule
\end{tabular}
\vspace{-0.5em}
\label{tab:reward_short}
\end{table}

\begin{figure}[t]
\centering
\includegraphics[width=0.48\textwidth]{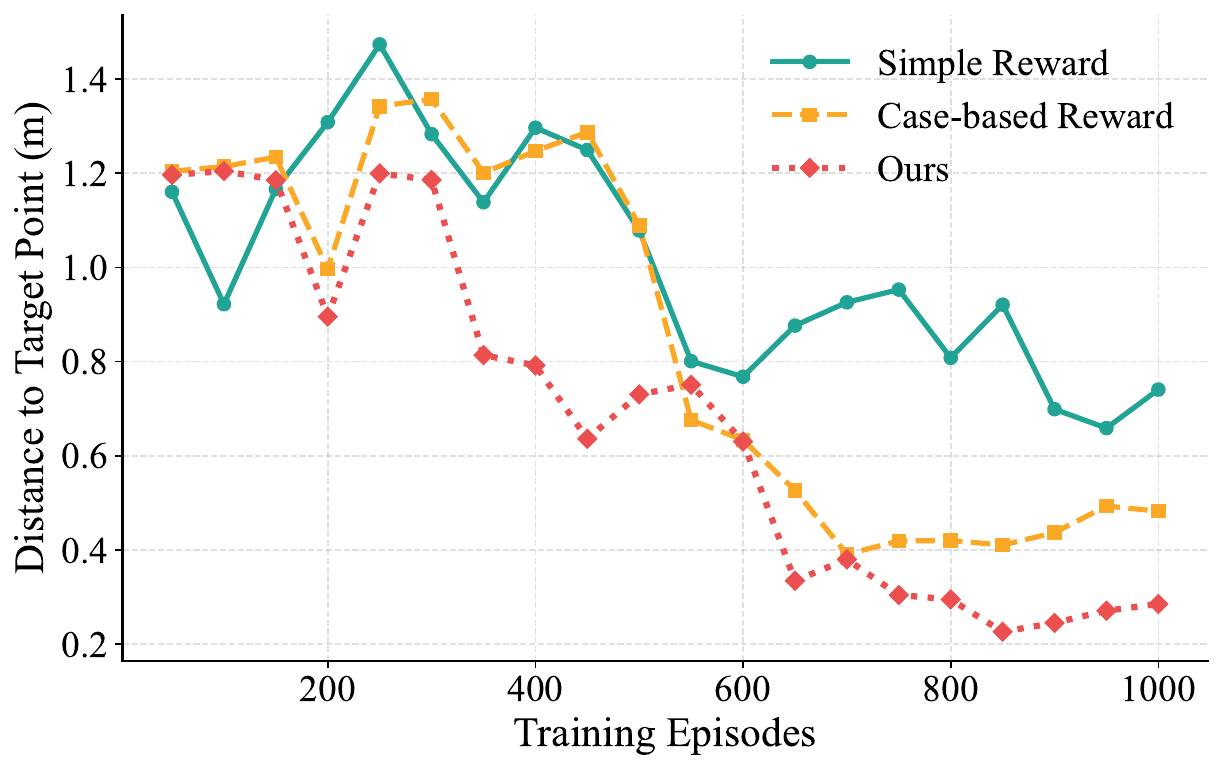}
\caption{\textbf{Learning Curves of Hitting Policies Trained with different reward designs over 1000 Iterations.}}
\label{fig:reward_comparison}
\end{figure}

To evaluate the effectiveness of the proposed \ac{CDTA} reward, we conduct experiments at a ball speed of 4 m/s with randomly sampled initial ball positions.
The experimental results are shown in the Fig. \ref{pic_all}. The figure reports the performance of all methods over 1000 training episodes. As training progresses, the proposed method demonstrates a substantially faster performance improvement, while the baseline methods converge more slowly and remain at higher error levels. This indicates that our approach can exploit training samples more effectively, leading to accelerated convergence and improved final performance. By the end of training, our method converges to the lowest return-to-target distance among all compared approaches, achieving an average error of approximately 226 mm, which corresponds to an improvement of at least 55.5\% over \ac{SOTA} methods.

Additionally, To validate the effectiveness of both the case-based structure and the temporally adaptive reward adjustment in the proposed \ac{CDTA} reward, we conducted an ablation study with two reward settings: a simple reward and a case-based reward.
The simple reward gives a reward only when the ball lands on the opponent's table. The case-based reward, compared with our method, lacks dynamic adjustment of reward weights throughout training. Table \ref{tab:reward_short} presents the successful return rates achieved by each method under different experimental settings and training episodes, where a successful return is defined as hitting the ball back onto the opponent’s table. The results indicate that our method attains higher return rates with fewer training samples. Fig. \ref{fig:reward_comparison} illustrates the variation in return accuracy under three different reward settings for a ball speed of 4 m/s. As the number of training episodes increases, our method reduces the distance between the ball landing point and the target more rapidly, resulting in more precise returns.

\subsubsection{Improved Convergence in High-speed Scenarios with the Reward-threshold Mechanism}
\begin{figure}[t]
\centering
\includegraphics[width=0.9\linewidth]{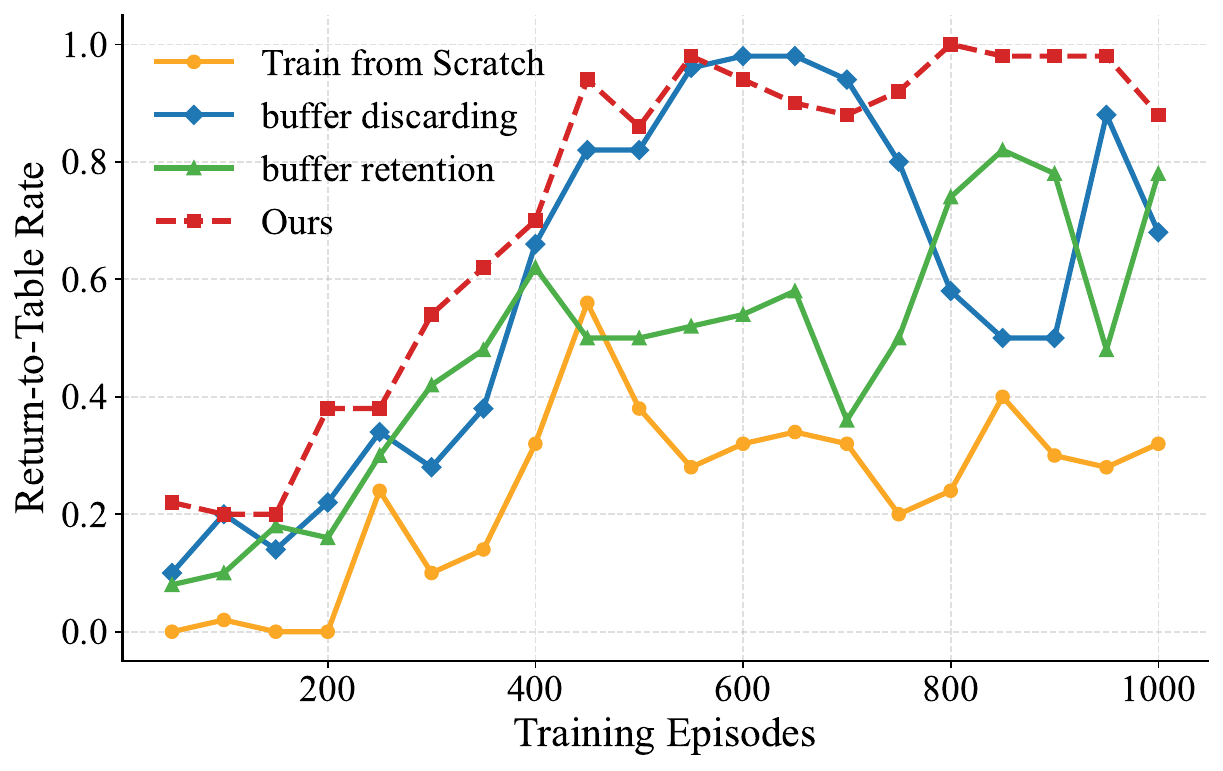}
\caption{\textbf{Return-to-table rate comparison under the high-speed scenario (5 m/s) between policies trained from scratch and those adapted from a pre-trained model.}}
\label{fig:reward_return}
\end{figure}

\begin{figure}[t]
\centering
\includegraphics[width=0.9\linewidth]{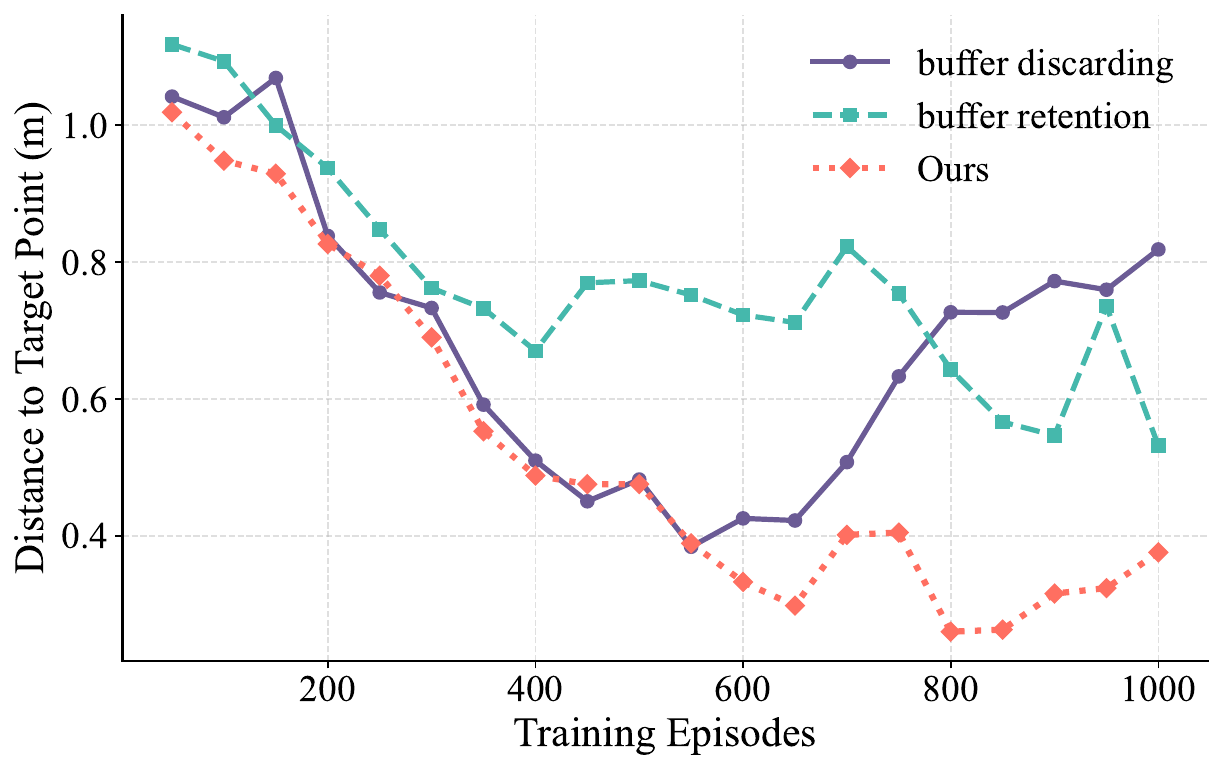}
\caption{\textbf{Comparison of target-point return errors for three models adapted from pre-training under a high-speed (5 m/s) scenario.}}
\label{fig:reward_buffer}
\end{figure}

To validate the effectiveness of the reward-threshold mechanism, we performed experiments in a high-speed scenario with a ball speed of 5 m/s. In these experiments, we compared a policy trained from scratch with policies adapted from a model pre-trained at a ball speed of 3~m/s.
For the models adapted from pre-training, three strategies were adopted to handle the transition of the replay buffer: 
(1) buffer discarding~\cite{fedus2020revisiting}, 
(2) buffer retention~\cite{isele2018selective}, and 
(3) buffer filtering using the proposed reward-threshold mechanism. 

The experimental results are shown in Fig.~\ref{fig:reward_return}. Training from scratch exhibits difficulty in converging in the higher speed setting, whereas leveraging a pre trained model improves convergence behavior. Furthermore, compared with other replay buffer handling strategies, our method yields more stable performance improvements and achieves a higher return rate.

We also evaluated the return accuracy of different strategies. As shown in Fig.~\ref{fig:reward_buffer}, our method exhibits a faster and more stable decrease in the distance between the ball landing point and the target, thereby achieving higher return accuracy than the other two strategies.

\section{CONCLUSIONS}
In this work, we present a biologically inspired approach for high-speed table tennis robots, combining event-based perception with sample-efficient learning. Specifically, we leverage the motion patterns and geometric features of the ball in event streams to perform ball detection in complex dynamic environments, significantly reducing the processed data and perception latency while improving detection accuracy. The effectiveness of this perception method is validated in real-world table tennis rally scenarios. Furthermore, we propose a human-inspired, sample-efficient training strategy that combines a \ac{CDTA} reward with a reward-threshold mechanism.
This strategy allows the agent to gradually adapt from low to high ball velocities, while progressively learning from basic to advanced skills at each velocity, emulating human learning processes.
Overall, this study demonstrates the potential of a biologically inspired approach that integrates event-based perception with sample-efficient \ac{RL} in a table tennis robot, providing a promising framework for high-speed dynamic tasks in real-world robotic applications.

\addtolength{\textheight}{-0cm} 

\bibliographystyle{IEEEtran}

\bibliography{reference}

\end{document}